\documentclass[nohyperref]{article}

\usepackage[round]{natbib}
\renewcommand{\bibname}{References}
\renewcommand{\bibsection}{\subsubsection*{\bibname}}

\usepackage{comment}
\usepackage{hyperref}

\usepackage{times}
\usepackage{color}
\usepackage{amsmath}
\usepackage{graphicx,psfrag,epsf}
\usepackage{enumerate}
\usepackage{url}
\usepackage{amssymb,amsfonts,mathtools}
\usepackage{bm}
\usepackage{color, colortbl}
\usepackage{algorithm}
\usepackage[ruled,vlined,algo2e, noend]{algorithm2e}
\usepackage{subcaption}

\usepackage{balance} 
\usepackage{multirow}
\usepackage{wrapfig,bbm}
\usepackage{bbm}
\usepackage{booktabs}

\usepackage[accepted]{arxiv}

\usepackage{amsmath}
\usepackage{amssymb}
\usepackage{mathtools}
\usepackage{amsthm}

\usepackage[capitalize,noabbrev]{cleveref}

\theoremstyle{plain}
\newtheorem{theorem}{Theorem}[section]
\newtheorem{proposition}[theorem]{Proposition}

\theoremstyle{definition}

\theoremstyle{remark}
\newtheorem{remark}[theorem]{Remark}

\SetKwInput{Input}{Input}
\SetKwProg{kwSystem}{System executes}{:}{}
\SetKwProg{kwServer}{ServerUpdate}{:}{}
\SetKwProg{kwClient}{ClientUpdate}{:}{}

\DeclarePairedDelimiterX{\norm}[1]{\lVert}{\rVert}{#1}
\DeclarePairedDelimiterX{\bnorm}[1]{\biggl\lVert}{\biggr\rVert}{#1}
\DeclarePairedDelimiterX{\abs}[1]{\lvert}{\rvert}{#1}

\def\de{\overset{\Delta}{=}} 
\def\var{\mathbb{V}} 
\def\E{\mathbb{E}} 

\def\T{{ \mathrm{\scriptscriptstyle T} }} 

\def\N{\mathcal{N}}

\def\vT{\sigma_{0}^2} 
\def\vD{\sigma^2} 
\def\ivT{\sigma_{0}^{-2}} 
\def\ivD{\sigma^{-2}} 
\def\th{\theta} 
\def\thH{\hat{\theta}}
\def\thL{\theta^{(\textsc{l})}} 
\def\thFL{\theta^{(\textsc{fl})}} 
\def\thG{\theta^{(\textsc{g})}} 
\def\vL{v^{(\textsc{l})}} 
\def\vFL{v^{(\textsc{fl})}} 
\def\vG{v^{(\textsc{g})}} 
\def\G{\textsc{gain}} 
\def\init{\textsc{init}} 
\def\initT{\th^{\init}} 
\def\iid{\textsc{iid}}

\def\limd{\rightarrow_{d}}
\def\lim{\rightarrow}
\def\w{w} 
\def\loss{\textrm{loss}}

\def\sumK{\sum_{k: \, k \neq m} w_k} 
\def\sumkA{\sum_{k}}
\def\summA{\sum_{m}}
\def\q{q}
\def\cov{\textrm{Cov}}
\def\cv{c_v} 
\def\wmin{w_{\textrm{min}}}
\def\wmax{w_{\textrm{max}}}

\newcommand{\sys}{Self-FL}

\icmltitlerunning{Self-Aware Personalized Federated Learning}

\begin{document}

\twocolumn[
\icmltitle{Self-Aware Personalized Federated Learning}




\begin{icmlauthorlist}
\icmlauthor{Huili Chen}{comp}
\icmlauthor{Jie Ding}{comp}
\icmlauthor{Eric Tramel}{comp}
\icmlauthor{Shuang Wu}{comp}
\icmlauthor{Anit Kumar Sahu}{comp}
\icmlauthor{Salman Avestimehr}{comp}
\icmlauthor{Tao Zhang}{comp}
\end{icmlauthorlist}

\icmlaffiliation{comp}{Alexa AI, Amazon, USA}

\icmlcorrespondingauthor{Huili Chen}{huc044@ucsd.edu}
\icmlcorrespondingauthor{Jie Ding}{jiedi@amazon.com}

\icmlkeywords{Machine Learning, ICML}

\vskip 0.3in
]



\printAffiliationsAndNotice{}  

\begin{abstract}
In the context of personalized federated learning~(FL), the critical challenge is to balance local model improvement and global model tuning when the personal and global objectives may not be exactly aligned.
Inspired by Bayesian hierarchical models, we develop a self-aware personalized FL method where each client can automatically balance the training of its local personal model and the global model that implicitly contributes to other clients' training. Such a balance is derived from the \textit{inter-client and intra-client uncertainty quantification}. A larger inter-client variation implies more personalization is needed. Correspondingly, our method uses \textit{uncertainty-driven local training steps and aggregation rule} instead of conventional local fine-tuning and sample size-based aggregation. With experimental studies on synthetic data, Amazon Alexa audio data, and public datasets such as MNIST, FEMNIST, CIFAR10 and Sent140, we show that our proposed method can achieve significantly improved personalization performance compared with the existing counterparts. 
\end{abstract}




\vspace{-2em}
\section{Introduction}
\label{sec_intro}
\vspace{-0.5em}

Federated learning (FL)~\citep{konevcny2016federated,mcmahan2017communication} is transforming machine learning (ML) ecosystems from ``centralized in-the-cloud'' to ``distributed across-clients,'' to potentially leverage the computation and data resources of billions of edge devices~\citep{lim2020federated}, without raw data leaving the devices. As a distributed ML framework, FL aims to train a global model that aggregates gradients or model updates from the participating edge devices. 
Recent research in FL has significantly extended its original scope to address the emerging concern of personalization, a broad term that often refers to an FL system that accommodates client-specific data distributions of interest~\citep{ding2022federated}.

\vspace{-0.3em}
In particular, each client in a personalized FL system holds data that can be potentially non-identically and independently distributed~(non-IID). 
For example, smart edge devices at different houses may collect audio data~\citep{purington2017alexa} of heterogeneous nature due to, e.g., accents, background noises, and house structures. Each device hopes to improve its on-device model through personalized FL without transmitting sensitive data.

\vspace{-0.3em}
While the practical benefits of personalization have been widely acknowledged, its theoretical understanding remains unclear. Existing works on personalized FL often derive algorithms based on a pre-specified optimization formulation, but explanations regarding the formulation or its tuning parameters rarely go beyond heuristic statements.

\vspace{-0.3em}
In this work, we take a different approach. Instead of specifying an optimization problem to solve, we start with a toy example and develop insights into the nature of personalization from a statistical uncertainty perspective. 
In particular, we aim to answer the following critical questions regarding personalized FL.

\vspace{-0.2em}
\textit{(Q1) It is natural to consider two extreme cases, namely, each client performs local training without FL, and all clients participate in conventional FL training. Both cases lead to practical lower-bound baselines of personalized FL performance of each client. But what would be an upper-bound baseline for the client? Or even theoretically, what would be the limit of improvement (in a proper sense) that a client could gain from participating in FL?}

\vspace{-0.2em}
\textit{(Q2) Suppose that the goal of each client is to improve its local model performance. How to design an FL training that leverages the server-side global model as an intermediary to exchange helpful information across clients? In particular, how to interpret the ideal global model and suitably aggregate local models to obtain it, and how to fine-tune each client's local training automatically?}

\vspace{-0.2em}
Both questions Q1 and Q2 are quite challenging.
Q1 demands a systematic way to characterize the client-specific and globally-shared information. Such a characterization is agnostic to any particular training process being used. To this end, instead of studying personalized data in full generality, we restrict our attention to a simplified and analytically tractable setting: \textit{two-level Bayesian hierarchical models}, where the top and bottom level describes inter-client and intra-client uncertainty, respectively. 

\vspace{-0.2em}
The above Q2 requires FL updates to be adaptive to the nature of the underlying discrepancy among client-specific local data, including sample sizes and distributions. The popular aggregation approach that uses a sample size-based weighting mechanism~\citep{mcmahan2017communication} does not account for the distribution discrepancy. Meanwhile, since clients' data is not shared, estimating the underlying data distributions is unrealistic. Consequently, it is highly nontrivial to measure and calibrate such a discrepancy in FL settings. Addressing the above issues is a key motivation of our work.

\vspace{-0.5em}
\subsection{Contributions}
\vspace{-0.2em}
The main contributions of this work are two-fold.

First, we propose to interpret personalization from a \textit{hierarchical model-based} perspective. 
We show that a central component towards understanding that limit lies in \textit{uncertainty quantification} of inter- and intra-client samples. Then, we develop a {Self-Aware Personalized FL (\sys{})} method that adaptively guides client-side training and server-side aggregation. 
The developed algorithm and its convergence are justified for two-level hierarchical models, which allow analytically tractable solutions and insights. 

Second, we propose a practical version of \sys{} for deep learning problems, which has the following advantages. Firstly, it can effectively enhance personalized performance on client-specific data, with the same level of computation and communication overhead for both the clients and the server, compared with the standard FedAvg algorithm~\citep{mcmahan2017communication}.
Secondly, compared with existing personalized FL methods that typically involve hyper-parameter tuning, it has \textit{built-in auto-tuning} for each client, which can significantly reduce nuisances for FL developers.  
We demonstrate the promising performance of \sys{} on synthetic data, public image and text data,
and private audio data from Amazon Alexa devices. Empirical results show that \sys{} achieves superior personalization performance compared with some popular methods. 

To our best knowledge, \sys{} is the first work that connects personalized FL with hierarchical modeling and utilizes uncertainty quantification to drive personalization.

\vspace{-0.2em}
\subsection{Related work}  \label{sec:related}
\vspace{-0.2em}

\textbf{Heterogeneous FL.} In earlier studies of FL, it was widely assumed that local models have to share the same architecture as the global model~\citep{li2020federated} to produce a single global inference model. This assumption limits the global model's complexity for the most data-indigent client. To personalize the computation and communication capabilities of each client, the work of~\citep{DingHeteroFL} proposed a new FL framework named {HeteroFL} to train heterogeneous local models and still produce a single global inference model. Although this {model heterogeneity} can be regarded as personalization of client-side resources, it differs significantly from personalized FL in this work, where our goal is to derive client-specific models due to clients' heterogeneous data. 
To address system heterogeneity, methods based on asynchronous communication and active client sampling have also been developed~\citep{bonawitz2019towards,nishio2019client}.  
To encourage a fair distribution of accuracy across clients, a method based on client-weighted loss was developed in~\cite{li2019fair}.

\vspace{-0.1em}

\textbf{Assisted learning.} Beyond edge computing, emerging real-world applications concern the collaboration among organizational learners such as research labs, government agencies, or companies. However, to avoid leaking useful and possibly proprietary information, an organization typically enforces stringent security measures, significantly limiting such collaboration. Assisted learning (AL)~\citep{DingAssist,DingGAL,diao2021privacy} has been developed as a decentralized framework for organizational learners (with rich data and computation resources) to autonomously assist each other without sharing data, task labels, or models. Consequently, each learner also obtains a ``personalized model'' to serve its own task. A critical difference between AL and FL is that participating learners in AL do not share task labels or a global model to meet organizational model privacy~\cite{DingIL}. Also, since a learner is resource-rich, AL often focuses on reducing the assistance rounds instead of communication costs at each round. As such, the techniques developed under AL are entirely different from those in the personalized FL literature. 

\vspace{-0.1em}
\textbf{Personalized FL.} The term \textit{personalization} in FL often refers to the development of client-specific model parameters for a given model architecture. In this context, each client aims to obtain a local model that has desirable test performance on its local data distribution. Personalized FL is critical for applications that involve statistical heterogeneity among clients. 

A research trend is to adapt the global model for accommodating personalized local models. To this end, prior works often integrate FL with other frameworks such as multi-task learning~\citep{smith2017federated},  meta-learning~\citep{jiang2019improving, khodak2019adaptive,fallah2020personalized,perfedavg,al2021data}, transfer learning~\citep{wang2019federated, mansour2020three}, knowledge distillation~\citep{li2019fedmd}, and lottery ticket hypothesis~\citep{li2020lotteryfl}. 
For example, DITTO~\citep{li2021ditto} formulates personalized FL as a multi-task learning (MTL) problem and regularizes the discrepancy of the local models to the global model using the $\ell_2$ norm.
FedEM~\cite{marfoq2021federated} uses the MTL formulation for personalization under a finite-mixture model assumption and provides federated EM-like algorithms.
We refer to \cite{vanhaesebrouck2017decentralized,hanzely2020lower,huang2021personalized} for more MTL-based approaches and theoretical bounds on the optimization problem.
From the perspective of meta-learning, pFedMe~\citep{pfedme} formulates a bi-level optimization problem for personalized FL and introduces Moreau envelopes as clients’ regularized loss. PerFedAvg~\citep{perfedavg} proposes a method to find a proper initial global model that allows a quick adaptation to local data. 
pFedHN~\cite{shamsian2021personalized} proposes to train a central hypernetwork for generating personalized models.  FedFOMO~\cite{zhang2020personalized} introduces a method to compute the optimally weighted model aggregation for personalization by characterizing the contribution of other models to one client.
Later on, we will focus on the experimental comparison of \sys{} with DITTO, pFedMe, and PerFedAvg, since they represent two popular personalized FL formulations, namely multi-task learning and meta-learning.

A limitation of existing personalized FL methods is that they do not provide a clear answer to question Q1. Consequently, it is unclear how to interpret the FL-trained results.
For example, in the extreme case where the clients are equipped with irrelevant tasks and data, any personalized FL methods that require a pre-specified global objective (e.g., Ditto~\citep{li2021ditto}, pFedMe~\citep{pfedme}) may cause significant biases to local clients.

\vspace{-0.5em}
\section{Bayesian Perspective of Personalized FL} \label{sec_form}
\vspace{-0.2em}

We discuss how \sys{} approaches personalized FL with theoretical insights from the Bayesian perspective in this section. The notations are defined as follows. 
Let $\N(\mu, \sigma^2)$ denote Gaussian distribution with mean $\mu$ and variance $\sigma^2$.
For a positive integer $M$, let $[M]$ denote the set $\{1,\ldots,M\}$. Let $\sum_{m \neq i}$ denote the summation over all $m\in [M]$ except for $m=i$. 
Some frequently used notations are summarized in Table~\ref{tab_notation} of the Appendix.

\vspace{-0.5em}
\subsection{Understanding personalized FL through a two-level Gaussian model} \label{subsec_toy}
\vspace{-0.3em}

To develop insights, we first restrict our attention to the following simplified case.
Suppose that there are $M$ clients. From the server's perspective, it is postulated that data $z_1,\ldots,z_M$ are generated from the following two-layer Bayesian hierarchical model: 
\begin{align}
    &\theta_m \mid \theta_0 \overset{\iid}{\sim} \N(\theta_0, \vT), \quad 
    z_{m} \mid \theta_m \overset{\iid}{\sim} \N(\theta_m, \vD_m), \label{eq4}
\end{align}
for all clients with indices $m=1,\ldots,M$. Here, $\vT$ is a constant, and $\theta_0$ is a hyperparameter with a non-informative flat prior (denoted by $\pi_0$). 
The above model represents both the connections and heterogeneity across clients. 
In particular, each client's data are distributed according to a client-specific parameter ($\theta_m$), which follows a distribution decided by a parent parameter ($\th_0)$. The parent parameter is interpreted as the root of shared information. 
In the rest of the section, we often study client 1's local model as parameterized by $\theta_1$ without loss of generality. 
Under the above model assumption, the parent parameter $\th_0$ that represents the global model has a posterior distribution $p(\theta_0 \mid z_{1:M}) \sim \N(\thG, \vG)$, where:
\vspace{-0.5em}
\begin{align} 
    &\thG \de \frac{\sum_{m \in [M]} (\vT + \vD_m)^{-1} z_m}{ \sum_{m \in [M]} (\vT + \vD_m)^{-1}}, \label{eq15}    \\ 
    &\vG \de \frac{1}{\sum_{m \in [M]} (\vT + \vD_m)^{-1}}. \nonumber 
\end{align}
\vspace{-0.8em}

The $z_m$ in Eqn.~(\ref{eq15}) may also be denoted by $\theta^{(L)}_m$, for reasons that will be seen in Eqn.~(\ref{eq1}).
From the perspective of client $m$, we suppose that the postulated model is Eqn.~(\ref{eq4}) for $m=2,\ldots,M$, and $\th_1=\th_0$.
It can be verified that the posterior distributions of $\th_1$ without and with global Bayesian learning are $ 
    p(\theta_1 \mid z_{1}) \sim \N(\thL_1, \vL_1)$ and $
    p(\theta_1 \mid z_{1:M}) \sim \N(\thFL_1, \vFL_1), 
$ 
respectively, which can be computed as: 
\begin{align}
    &\thL_1 \de z_1 , \quad  \vL_1 \de \vD_1 , \nonumber \\
    &\thFL_1 \de
    \frac{\ivD_1 \thL_1 + \sum_{m\neq 1} (\vT + \vD_m)^{-1} \thL_m}{ \ivD_1 + \sum_{m\neq 1} (\vT + \vD_m)^{-1}}, \label{eq1}   \\
    &\vFL_1 \de \frac{1}{\ivD_1 + \sum_{m\neq 1} (\vT + \vD_m)^{-1}}. \nonumber  
\end{align}
The first distribution above describes the learned result of client 1 from its local data, while the second one represents the knowledge from all the clients' data in hindsight. 
Using the mean square error as risk, the Bayes estimate of $\theta_1$ or $\theta_0$ is the mean of the posterior distribution, namely $\thL_1$ and $\thFL_1$ as defined above.


The flat prior on $\th_0$ can be replaced with any other distribution to bake prior knowledge into the calculation. We consider the flat prior because the knowledge of the shared model is often vague in practice.
The above posterior mean $\thFL_1$ can be regarded as the optimal point estimation of $\theta_1$ given all the clients' data, thus is referred to as ``FL-optimal''. $\thG$ can be regarded as the ``global-optimal.'' The posterior variance quantifies the reduced uncertainty conditional on other clients' data. Specifically, we define the following \textit{Personalized FL gain} for client 1 as: 
\vspace{-0.5em}
\begin{align}
   \G_1 
   &\de \frac{\vL_1}{\vFL_1}   
   = 1 + \vD_1  \sum_{m\neq 1} (\vT + \vD_m)^{-1}. \nonumber
\end{align}

\vspace{-0.5em}
\begin{remark}[Interpretations of the posterior quantities] \label{remark:interpret_post}
Each client, say client 1, aims to learn $\th_1$ in the personalized FL context. Its learned information regarding $\th_1$ is represented by the Bayesian posterior of $\th_1$ conditional on either its local data $z_1$ (without communications with others), or the data $z_{1:M}$ in hindsight (with communications). 
For the former case, the posterior uncertainty described by $\vL_1$ depends only on the local data quality $\vD_1$.
For the latter case, the posterior mean $\thFL_1$ is a weighted sum of clients' local posterior means, and the uncertainty will be reduced by a factor of $\G_1$.
Since a point estimation of $\th_1$ is of particular interest in practical implementations, we treat $\thFL_1$ as the theoretical limit in the FL context (recall question Q1).
To develop further insights into $\thFL_1$, we consider the following extreme cases. 

\vspace{-0.3em}
$\bullet$ As $\vT \rightarrow \infty$, meaning that the clients are barely connected, the quantities $\thFL_1$ and $\vFL_1$ reduce to $\thL_1$ and $\vL_1$, respectively; meanwhile, the personalized FL gain approaches one, the global parameter mean $\thG$ becomes a simple average of $\thL_m$ (or $z_m$), and the global parameter has a large variance. 

$\bullet$ When $\vT = 0$, meaning that the clients follow the same underlying data-generating process and the personalized FL becomes a standard FL, we have $\thFL_1 = \thG$, which is a weighted sum of clients' local optimal solution with weight proportional to  $\ivD_m$ (namely client $m$'s precision). 

$\bullet$ When $\vD_1$ is much smaller than all other $\vD_m$'s and $\vT$, and $M$ is not too large, meaning that client 1 has much higher quality data compared with the other clients combined, we have $\thFL_1 \approx z_1 = \thL_1$ and $\G_1 \approx 1$. In other words, client 1 almost learns on its own.
Meanwhile, client 1 can still contribute to other clients through the globally shared parameter $\th_0$. For example, the gain for client 2 would be 
$\G_2 \geq \vD_2/ \vD_1 $, which is much larger than one.
\end{remark}

\vspace{0.2em}
\begin{remark}[Local training steps to achieve $\thFL_1$] 
Suppose that client 1 performs $\ell$ training steps using its local data and negative log-likelihood loss. 
We show that with a suitable number of steps and initial value, client 1 can obtain the intended $\thFL_1$.
The local objective is: 
\begin{align}
    \th \mapsto (\th- z_1)^2 / (2 \vD_1)= (\th- \thL_1)^2 / (2 \vD_1), \label{eq9}
\end{align}
which coincides with the quadratic loss.
Let $\eta \in (0,1)$ denote the learning rate. 
By running the gradient descent:  
\begin{align}
    \th^{\ell}_1 &\leftarrow \th^{\ell-1}_1 - \eta  \frac{\partial }{\partial \theta} \biggl( (\th- \thL_1)^2 / (2 \vD_1) \biggr)|_{\th^{\ell-1}_1} \nonumber \\
    &= \th^{\ell-1}_1 - \eta (\th^{\ell-1}_1- \thL_1) / \vD_1
\end{align}
for $\ell$ steps with the initial value $\initT_1$, client 1 will obtain: 
\begin{align}
    \th^{\ell}_1 = \bigl( 1 - (1-\ivD_1 \eta)^\ell \bigr) \thL_1 + (1-\ivD_1 \eta)^\ell \initT_1 . \label{eq2}
\end{align}
It can be verified that Eqn.~(\ref{eq2}) becomes $\thFL_1$ in Eqn.~(\ref{eq1}) if and only if:  
\begin{align}
    \initT_1 
    & =\frac{ \sum_{m\neq 1} (\vT + \vD_m)^{-1} \thL_m }{ \sum_{m\neq 1} (\vT + \vD_m)^{-1} } , \label{eq_08}  \\
    (1-\ivD_1 \eta)^\ell 
    &=  \frac{ \sum_{m\neq 1} (\vT + \vD_m)^{-1}}{ \ivD_1 + \sum_{m\neq 1} (\vT + \vD_m)^{-1} }. \label{eq_07}  
\end{align}
In other words, with a suitably chosen initial value $\initT_1$, learning rate $\eta$, and the number of (early-stop) steps $\ell$, client 1 can obtain the desired $\thFL_1$.
\end{remark}


\vspace{-0.6em}
\subsection{General parametric models} \label{subsec_param}
\vspace{-0.5em}

In a more general parametric model setting, the nature of the personalized FL problem remains the same as the simple case in Subsection~\ref{subsec_toy}. 
Suppose that there are $M$ clients, and their data are assumed to be generated from: 
\begin{align}
    \theta_m \mid \theta_0 \overset{\iid}{\sim} \N(\theta_0, \vT), \, 
     \w_{m,i} \overset{\iid}{\sim}  p_m(\cdot \mid \theta_m),  i \in [N_m]. \nonumber 
\end{align}
For simplicity, we suppose that each observation $\w_{m, i}$ is a scalar. The related discussions can be easily extended to multivariate settings. 
Compared with Eqn.~(\ref{eq4}), each client may have a different amount of data and follow client-specific distributions (denoted by $p_m$). 
Let $z_m \de \thL_m$ denote the minimum of the negative log-likelihood function
$L_m(\th_m) = \sum_{i=1}^{N_m} \loss (\w_{m,i}, \th_m)$.
Standard asymptotic statistics for $M$-estimators~\citep[Ch.5]{van2000asymptotic} under regularity conditions show that 
$\sqrt{N_m} (z_{m} - \theta_m) \limd \N(0, v_m^2)$ as $N_m \rightarrow \infty$, with a constant $v_m^2$ (the inverse Fisher information). Thus, we approximately have: 
\begin{align}
    z_{m} \mid \theta_m &\sim \N(\theta_m, N_m^{-1} v_m^2). \nonumber
\end{align}
In other words, we may treat the statistic $\theta^{(L)}_m$ as the ``data'' $z_m$ in the two-level Gaussian model in Subsections~\ref{subsec_toy}.
Letting $\vD_m \de N_m^{-1} v_m^2$ and taking it into the equations in Subsection~\ref{subsec_toy}, we can approximate posterior quantities accordingly. Also, the objective function $L_m(\th_m)$ is approximated by its second-order Taylor expansion in the form of Eqn.~(\ref{eq9}).
It is worth mentioning that the asymptotic results for parametric models may not hold for neural networks. 

In particular, we let:
\vspace{-0.3em}
{\small 
\begin{align}
     &\thFL_1 \de 
    \frac{N_1 v^{-2}_1 \thL_1 + \sum_{m\neq 1} (\vT + N_m^{-1}v_m^2)^{-1} \thL_m}{ N_1 v^{-2}_1 + \sum_{m\neq 1} (\vT + N_m^{-1} v_m^2)^{-1} } , \nonumber \\
    &\thG \de \frac{\sum_{m \in [M]} (\vT + N_m^{-1} v_m^2)^{-1} \thL_m}{ \sum_{m \in [M]} (\vT + N_m^{-1} v_m^2)^{-1} } . \nonumber 
\end{align}
}
Each client, say client 1, can obtain its personal optimal solution $\thFL_1$ through the following initial value and optimization parameters $\eta, \ell$.
\vspace{-0.3em}
{\small 
\begin{align}
    \initT_1 &=\frac{ \sum_{m\neq 1} (\vT + N_m^{-1} v_m^2)^{-1} \thL_m }{ \sum_{m\neq 1} (\vT + N_m^{-1} v_m^2)^{-1} }, \label{eq_8}   \\
    (1- N_1 v_1^{-2} \eta)^\ell &=  \frac{ \sum_{m\neq 1} (\vT + N_m^{-1} v_m^2)^{-1}}{ N_1  v^{-2}_1 + \sum_{m\neq 1} (\vT + N_m^{-1} v_m^2)^{-1} }.
    \label{eq7} 
\end{align}
}

\begin{remark}[Interpretations of the choice of $\eta$ and $\ell$]\label{remark_opt}
Let us consider the following extreme cases. 

$\bullet$ (Large sample size, few clients) Suppose that $N_1=\cdots=N_M$ $\gg M > 1$, and $v^2_1, \vT$ are comparable. Then, for client 1, Eqn.~(\ref{eq7}) becomes:
\vspace{-0.3em}
{\small
\begin{align}
    (1- N_1 v_1^{-2} \eta)^\ell \approx \frac{(M-1) \ivT}{N_1 v^{-2}_1 + (M-1) \ivT} \approx \frac{M}{N_1} \frac{v^2_1}{\vT}, \nonumber
\end{align}
}
which implies that client 1 needs to run aggressively with its local data. Also, the smaller $\sigma_1^2=v_1^2/N_1$ (namely better data quality or larger sample size), the more local training steps is favored.

$\bullet$ (Small sample size, many clients) Suppose that $1 \ll N_1=\cdots=N_M \ll M$, and $v^2_1=\cdots = v^2_M$. We have:  
{\small
\begin{align}
    (1- N_1 v_1^{-2} \eta)^\ell &\approx \frac{(M-1)  (\vT + N_1^{-1} v_1^2)^{-1} }{N_1 v^{-2}_1 + (M-1)  (\vT + N_1^{-1} v_1^2)^{-1}} \nonumber  \\  
    &\approx 1-  \frac{N_1}{M} \frac{\vT }{v^2_1},
\end{align}
}

which implies that client 1 needs to set: 
\begin{align}
    N_1 v_1^{-2} \eta \cdot \ell \approx \frac{N_1}{M} \frac{\vT }{v^2_1}, \quad \textrm{ or } \quad
    \eta \cdot \ell \approx \frac{\vT}{M} . \nonumber
\end{align}
It is interesting to see that the choice of $\eta, \ell$ is independent of the client in this scenario. 

$\bullet$ (Homogeneous clients) Suppose that $\vT=0$, meaning no discrepancy between clients' data distributions. We have
\begin{align}
    (1- N_1 v_1^{-2} \eta)^\ell = 
    \frac{ \sum_{m\neq 1} N_m v^{-2}_m }{ \sum_{m\in [M]} N_m v^{-2}_m }. \nonumber
\end{align}
If we further assume that $v_1=\cdots=v_M$, $M$ is large, and $N_1,\ldots,N_M$ are at the same order, client~1 needs to set 
$
    N_1 v_1^{-2} \eta \cdot \ell \approx N_1/N,
$
where $N \de N_1+\cdots+N_M$. In other words, $\eta \cdot \ell \approx v_1^{2}/N$, which is the same among all clients.
\end{remark}

\section{Proposed Solution for Personalized FL} \label{sec_method}

{Our proposed \sys{} framework has three key components as detailed in this section: (i) proper initialization for local clients at each round, (ii) automatic determination of the local training steps, (iii) discrepancy-aware aggregation rule for the global model. These components are interconnected and contribute together to \sys{}'s effectiveness. Note that points (i) and (iii) direct \sys{} to the regions that benefit personalization in the optimization space during local training, which is not considered in prior works such as DITTO~\cite{li2021ditto} and pFedMe~\cite{pfedme}. Therefore, \sys{} is more than imposing implicit regularization via early stopping.}

\vspace{-0.3em}
\subsection{From posterior quantities to FL updating rules} \label{subsec_derive}
\vspace{-0.2em}
In this section, we show how the posterior quantities of interest in Section~\ref{sec_form} can be connected with FL, where clients' parameters suitably deviate from the global parameter, and the global parameter is a proper aggregation of clients' parameters.
For notional brevity, we still consider the two-level Gaussian model in Subsection~\ref{subsec_toy}. For regular parametric models, we mentioned in Subsection~\ref{subsec_param} that one may treat $z_m$ as the local optimal solution $\thL_m$, and its variance $\vD_m$ as the variance of $\thL_m$ due to finite sample.


Recall that each client $m$ can obtain the FL-optimal solution $\thFL_m$ with the initial value $\initT_m$ in Eqn.~(\ref{eq_08}) and tuning parameters $\eta, \ell$ in Eqn.~(\ref{eq_07}). Also, it can be shown that $\initT_m$ is connected with the global-optimal $\thG$ defined in Eqn.~(\ref{eq15}) through: 
\begin{align}
    \initT_m = \thG -  \frac{(\vT + \vD_m)^{-1} }{\sum_{k: \, k \neq m} (\vT + \vD_k)^{-1}} (\thL_m - \thG). \label{eq12}
\end{align}
The initial value $\initT_m$ in Eqn.~(\ref{eq12}) is unknown during training since $\thL_m, \thG$ are both unknown. A natural solution is to update $\initT_m$, $\thL_m$, and $\thG$ iteratively, leading to the following personalized FL rule.

\textbf{Generic \sys{}}:
At the $t$-th ($t=1,2,\ldots$) round of FL:

\indent $\bullet$ \textit{Each client $m$} receives the latest global model $\th^{t-1}$ from the server (with an initial value $\th^0$), and calculates
{\small   
\begin{align}
    \th^{t,\init}_m \de \th^{t-1} -  \frac{(\vT + \vD_m)^{-1} }{\sum_{k: \, k \neq m} (\vT + \vD_k)^{-1}} (\th^{t-1}_m - \th^{t-1}) ,\label{eq_101}
\end{align}
}
where $ \th^{t-1}_m$ is client $m$'s latest personal parameter at round $t-1$, initialized to be $\th^0$.
Starting from the above $\th^{t,\init}_m$, client $m$ performs gradient descent-based local updates with optimization parameters following Eqn.~(\ref{eq_07}) or its approximations, and obtains a personal parameter $\th^{t}_m$.

\indent $\bullet$ \textit{Server} collects $\th^{t}_m$, $m=1,\ldots,M$, and calculates: 
\begin{align}
    \th^t &\de \frac{\sum_{m \in [M]} (\vT + \vD_m)^{-1} \th^{t}_m}{ \sum_{m \in [M]} (\vT + \vD_m)^{-1} } . \label{eq_102}
\end{align}
%
In general, the above $\vT, \vD_m$ represent ``inter-client uncertainty'' and ``intra-client uncertainty,'' respectively. When $\vT$ and $\vD_m$'s are unknown, they can be approximated using asymptotics of M-estimators or using practical finite-sample approximations (elaborated in Subsection~\ref{subsec_algo}).




We provide a theoretical understanding of the convergence. Consider the data-generating process that $\thL_m \sim \mathcal{N}(\theta_m, \vD_m)$ are independent, and $\theta_m \overset{\iid}{\sim} \N(\theta_0, \vT)$ where $\vT$ is a fixed constant and each $\vD_m$ may or may not depend on $M$. 
Let $O_p$ denote the standard stochastic boundedness. The following result gives an error bound of each client's personalized parameter and the server's parameter. 

\vspace{0.2em}
\begin{proposition} \label{prop_converge}
    Assume that $\max_{m\in [M]} \vD_m$ is upper bounded by a constant, and there exists a constant $\q \in (0,1)$ such that: 
\begin{align}
   \max_{m\in [M]} \frac{\sum_{k:\, k\neq m} (\vD_k + \vT)^{-1}}{\ivD_m + \sum_{k:\, k\neq m} (\vD_k + \vT)^{-1}} \leq \q . \label{eq103}
\end{align}
Suppose that at $t=0$, the gap between the initial parameter and each client's FL-optimal value satisfies $|\hat{\th}^{0} - \thFL_m| \leq C$ for all $m\in [M]$.
Then, for every positive integer $t$, the quantities
$
    \max_{m\in [1:M]} |\th^{t}_m - \thFL_m|, |\th^t - \thG|$ are both upper bounded by $C \cdot q^{t} + O_p(M^{-1/2})
$
as $M \rightarrow \infty$.

\end{proposition}

\vspace{0.3em}
\begin{remark} [Interpretation of Proposition~\ref{prop_converge}]
    The proposition shows that the estimation error of each personalized parameter $\th^{t}_m$ and the server parameter $\th^t$ can uniformly go to zero as the number of clients $M$ and the FL round $t$ go to infinity. The error bound involves two terms. The first term $q^t$ corresponds to the optimization error. Typically, every $\sigma_m$ is a small value. If each client has a sample size of $N$, $q$ will be at the order of $M/(N+M)$. The second term $O_p(M^{-1/2})$ is a statistical error that vanishes as more clients participate. Intuitively, the initial model of each client and the global model at each round are averaged over many clients, so a larger pool leads to a smaller bias. The proof is nontrivial because the averaged terms are statistically dependent. For the particular case that each $\vD_m$ is at the order of $N^{-1}$, we can see that the error is tiny when both $M$ and $N/M$ grow.
\end{remark}

\vspace{-0.5em}
\subsection{SGD-based practical algorithm for deep learning}\label{subsec_algo}
\vspace{-0.2em}

For the training method proposed in Subsection~\ref{subsec_derive}, the quantities $\vT$ and $\vD_m$ are crucial as they affect the choice of learning rate $\eta_m$ and the early-stop rule.
For many complex learning models, we do not know $\vT$ and $\vD_m$, and the asymptotic approximation of $\vT$ and $\vD_m$'s may not be valid due to a lack of regularity conditions.
Furthermore, one often uses SGD instead of GD to perform local updates, so the choice of the stop rule will depend on the batch size.
In this subsection, we consider the above aspects and develop a practical algorithm for deep learning.
Following the discussions in Subection~\ref{subsec_param}, we can generally treat $\vD_m$ as ``uncertainty of the local optimal solution $\thL_m$ of client $m$'', and $\vT$ as ``uncertainty of clients' underlying parameters.''
We propose a way to approximate them.

Assume that for each client $m$, we had $u$ independent samples of its data and the corresponding local optimal parameter $\th_{m,1},\ldots,\th_{m,u}$. We could then estimate $\vD_m$ by their sample variance. In practice, we can treat each round of local training as the optimization from a bootstrapped sample. In other words, at the round $t$, let client $m$ run sufficiently many steps (not limited by our tuning parameter $\ell$) until it approximately converges to a local minimum, denoted by $\th_{m,t}$. 
To save computational costs, a client does not necessarily wait for its local convergence. Instead, we recommend that each client use its personal parameter as a surrogate of $\th_{m,t}$ at each round $t$, namely to use $\th_m^t$.  
In other words, at round $t$, we approximate $\vD_m$ with:
\vspace{-0.6em}
\begin{align}
    \widehat{\vD_m} = \textrm{empirical variance of }\{\th_{m}^1, \ldots , \th_{m}^t \} .\label{eq_103}
\end{align}
\vspace{-0.6em}
Likewise, at round $t$, we estimate $\vT$ by:
\vspace{-0.2em}
\begin{align}
    \widehat{\vT} = \textrm{empirical variance of }\{\th_{1}^t,\ldots, \th_{M}^t \} . \label{eq_104}
    \vspace{-2.2em}
\end{align}

For multi-dimensional parameters, a counterpart of the derivations in earlier sections will involve matrix multiplications, which does not fit the usual SGD-based learning process. We thus simplify the problem by introducing the following uncertainty measures.  
For vectors $x_1, \ldots, x_M$, their empirical variance is defined as the trace of
$ 
  \sum_{m \in [M]} (x_m - \bar{x}) (x_m - \bar{x})^\T, 
$ 
which is the sum of entry-wise empirical variances.
$\widehat{\vD_m}$ and $\widehat{\vT}$ will be defined from such empirical variances, similar to Eqn.~(\ref{eq_103}) and~(\ref{eq_104}). 
In practice, for large neural network models, we suggest deploying \sys{} after the baseline FL training runs for certain rounds (namely warm-start) so that the fluctuation of local models does not hinder variance approximation.

\vspace{-0.3em}
Combining the above discussion and the generic method in Subsection~\ref{subsec_derive}, we propose our personalized FL solution in Algorithm~\ref{algo}.
We include the implementation details of \sys{} 
in Section~\ref{sec_exp} and Appendix.
{We derived how to learn $\thG$ and $\thFL_m$ for parametric models in Subsection~\ref{subsec_param}.
The key motivation of Algorithm~\ref{algo} is that a client often cannot obtain the exact $\thL_m$ (especially in DL), so we interweave local training and personalization through an iterative FL procedure. Note that the computation of $\thG$ and $\thFL_m$ requires uncertainty measurements approximated from communications between the clients and the server.}

\vspace{-0.5em}
\setlength{\textfloatsep}{8pt} 
\begin{algorithm}[!ht]
\SetAlgoLined
\DontPrintSemicolon
\caption{Self-Aware Personal FL (Self-FL)}
\label{algo}
\Input{A server and $M$ clients. System-wide objects, including the initial model parameters $\th_{m}^{0} = \th^0$, initial uncertainty of clients' parameters $\widehat{\vT}=0$, the activity rate $C$, the number of communication rounds $T$, server batch size $B_s$, and client batch size $B_m$, and the loss function $(z, \th) \mapsto \loss(z, \th)$ corresponding to a common model architecture. Each client $m$'s local resources, including $D_m$ that consists of $N_m$ data observations, parameter $\th_m$, number of local training steps $\ell_m$, and learning rate $\eta_m$.  
}
\kwSystem{}{
    \For{\textup{each communication round }$t = 1,\dots T$}{
        $\mathbb{M}_t \leftarrow \max(\lfloor C \cdot M \rfloor ,1)$ active clients uniformly sampled without replacement
    
        \For{\textup{each client }$m \in \mathbb{M}_t$\textup{ \textbf{in parallel}}}{
            Distribute server model parameters $\th^{t-1}$ to local client $m$ 
            
            $(\th_{m}^{t}, \widehat{\vD_m}) \leftarrow$ \textbf{ClientUpdate}$(D_m, \th^{t-1}, \widehat{\vT})$
        }
        $(\th^t, \widehat{\vT}) \leftarrow$ \textbf{ServerUpdate}$(\th_{m}^{t}, \widehat{\vD_m}, m \in \mathbb{M}_t)$
    }
}
\kwClient{\textup{$(D_m, \th^{t-1}, \widehat{\vT})$}}{
    Update inter-client uncertainty $\widehat{\vD_m}$ 
    using (\ref{eq_103})
    
    Calculate the initial parameter for local training:
    \vspace{-0.8em}
    \begin{align}
    \resizebox{.85\hsize}{!}{$\th_{m}^{t} \de \th^{t-1} -  \frac{(\widehat{\vT} + \widehat{\vD_m})^{-1} }{\sum_{k: \, k \neq m} (\widehat{\vT} + \widehat{\vD_k})^{-1}} (\th_{m, t-1} - \th^{t-1})$}
    \end{align}
    
    \vspace{-0.5em}
    Choose $\ell_m \geq 1$ according to:
    \vspace{-1em}
    \begin{align} 
    \label{eq:lm}
    \resizebox{.85\hsize}{!}{$\biggl(1-\frac{\eta_m }{B_m \widehat{\vD_m}} \biggr)^{\ell_m} = \frac{\sum_{k: \, k \neq m} (\widehat{\vT} + \widehat{\vD_k})^{-1}}{ 1/\widehat{\vD_m} + \sum_{k: \, k \neq m} (\widehat{\vT} + \widehat{\vD_k})^{-1}}$}
    \end{align}
    
    \vspace{-0.3em}
    \For{\textup{local step $s$ from $1$ to $\ell_m$}}{
        Sample a batch $I_s \subset [N_m]$ of size $B_m$\\
        Update $\th_{m}^{t} \leftarrow \th_{m}^{t} - \eta_m \nabla_{\th} \sum_{i \in I_s} \loss (\w_{m,i}, \th_{m}^{t}), $
    }
    Return $(\th_{m}^{t}, \widehat{\vD_m})$ and send them to the server
}
\kwServer{\textup{$(\th_{m}^{t}, \widehat{\vD_m}, m \in \mathbb{M}_t)$}}{
    \vspace{0.2em}
    Update inter-client uncertainty $\widehat{\vT}$ using the empirical variance of 
    $\{\th_{m}^{t}, m \in \mathbb{M}_t\}$.

    Calculate model parameters:
    \vspace{-1em}
    \begin{align}
        \th^t &\de \frac{\sum_{m \in \mathbb{M}_t} (\widehat{\vT} + \widehat{\vD_m})^{-1} \th^{t}_m}{ \sum_{m \in \mathbb{M}_t} (\widehat{\vT} + \widehat{\vD_m})^{-1} } . \label{eq_105}
    \end{align}
    \vspace{-1em}
    
    Return $(\th^t, \widehat{\vT})$
}
\end{algorithm} 

\section{Experimental Studies} \label{sec_exp}

\vspace{-0.3em}
\subsection{Experimental setup}  \label{sec:setup}
\vspace{-0.3em}
We use AWS p316 instances for all experiments. To evaluate the performance of \sys{}, we consider synthetic data, images, texts, and audios. 
{The loss in Alg.~\ref{algo} can be general. For our experiments on MNIST, FEMNIST, CIFAR10, Sent140, and wake-word detection, we used the cross-entropy loss.}
Our empirical results on DL problems suggest that \sys{} provides promising personalization performance in complex scenarios. 

\noindent \textbf{Synthetic Data.} We construct a synthetic dataset based on the two-layer Bayesian model discussed in Subsection~\ref{subsec_toy} where each data point is a scalar. 
The FL system has a total of $M=20$ clients, and the task is to estimate the global and local model parameters. 
The heterogeneity level of the data can be controlled by specifying the inter-client uncertainty $\sigma^2_{0}$ and the local dataset size $N_m$ for each client $m$. 
We construct two synthetic FL datasets, one with high homogeneity and one with high heterogeneity, to investigate the performance of FL algorithms in different scenarios. The empirical results are provided in {Appendix Subsection~\ref{sec:eval_synthetic}.}

\noindent \textbf{MNIST Data.} This image dataset has $10$ output classes and input dimension of $28 \times 28$. We use a multinomial logistic regression model for this classification task. The FL system has a total of $1000$ clients. We use the same non-IID MNIST dataset as FedProx~\citep{fedprox}, where each client has samples of two-digit classes, and the local sample size of clients follows a power law~\citep{li2018federated}.


\noindent \textbf{FEMNIST Data.} The federated extended MNIST (FEMNIST) dataset has $62$ output classes. The FL system has $M=200$ clients. We use the same approach as FedProx~\citep{li2018federated} for heterogeneous data generation where ten lower-case characters (`a'-`j') from EMNIST are selected, and each client holds five classes of them.


\noindent \textbf{CIFAR10 Data.} This image dataset has images of dimension $32 \times 32 \times 3$ and $10$ output classes. We use the non-i.i.d. data provided by the pFedMe repository~\cite{pfedme_repo}. There are a total of $20$ clients in the system where each user has images of three classes.

\noindent \textbf{Sent140 Data.} 
This is a text sentiment analysis dataset containing tweets from Sentiment140~\cite{go2009twitter}. It has two output classes and $772$ clients. We generate non-i.i.d. data using the same procedure as FedProx~\cite{fedprox}. Detailed setup and our experimental results on Sent140 are provided in Appendix Subsection~\ref{appendix:eval_sent140}.

\noindent \textbf{Private Wake-word Data.} We also evaluate \sys{} on a private audio data from Amazon Alexa for the wake-word detection task. This is a common task for smart home devices where the on-device model needs to determine whether the user says a pre-specified keyword to wake up the device. This task has two output classes, namely `wake-word detected' and `wake-word absent.' The audio stream from the speaker is represented as a spectrogram and is fed into a convolution neural network (CNN). 
The CNN is a stack of convolutional and max-pooling layers ($11$ in total). The number of trainable parameters is around $3\cdot 10^6$. 
The heterogeneity between clients comes from the intrinsic property of different device types (e.g., determined by hardware and use scenarios). 
We use an in-house far-field corpus that contains de-identified far-field data collected from millions of speakers under different conditions with users' content. 
This dataset contains $39$ thousand hours of training data and $14$ thousand hours of test data.

To support practical FL systems where only a subset of clients participate in one communication round (also known as client sampling), \sys{} can update the global model $\theta^t$ with \textit{smoothing average}. 
Particularly, given an activity rate (i.e., ratio of clients selected in each round) $C \in (0,1)$, we first obtain the current estimation $\thH^t$ by applying Eqn.~(\ref{eq_105}) to the active clients. Then, the server updates the global model using $\theta^t = (1-C) \cdot \theta^{t-1} + C \cdot \thH^t.$
Updating the global model with a smoothing average allows us to integrate historical information and reduce instability, especially for a small activity rate.

\vspace{-0.5em}
\subsection{Effectiveness on the image domain} \label{sec:eval_images}
\vspace{-0.3em}

\noindent \textbf{Results on MNIST and FEMNIST.} A description of MNIST and FEMNIST data are detailed in Subsection~\ref{sec:setup}. The activity rate is set to $C=0.1$.
For MNIST, we use learning rate $\eta=0.03$ and batch size $10$ as suggested in FedProx~\citep{fedprox}. 
For FEMNIST, we use $\eta=0.01$ and batch size $10$. 
The FL training starts from scratch and runs for $200$ rounds. For comparison, we implement FedAvg and three other personalized FL techniques, DITTO~\citep{li2021ditto}, pFedMe~\citep{pfedme}, and PerFedAvg~\citep{perfedavg} with the same hyper-parameter configurations.

\vspace{-0.2em}
For real-world data, the theoretical value of the local training steps $l_m$ computed using Eqn.~(\ref{eq:lm}) might be too large for the client (due to limitation of client's computation/memory budget). In this experiment, we determine the actual local training steps for \sys{} framework as $\hat{l}_m = \min\{l_m, l_{\textrm{max}} \}$ where $l_{\textrm{max}}$ is a pre-defined cutoff threshold. 
We use a fixed local training step of $20$ for other FL algorithms and set $l_{\textrm{max}}=40$ for \sys{}. 
Note that $\sigma^2_m$ represents the variance of $\theta_m$ (i.e., model parameters such as neural coefficients). In our experiments, $\sigma^2_m$ is approximated using the empirical variance of $\theta_m$ across rounds.

\begin{figure}[tb]
\begin{subfigure}[b]{0.49\columnwidth}
  \centering
  \includegraphics[width=\linewidth]{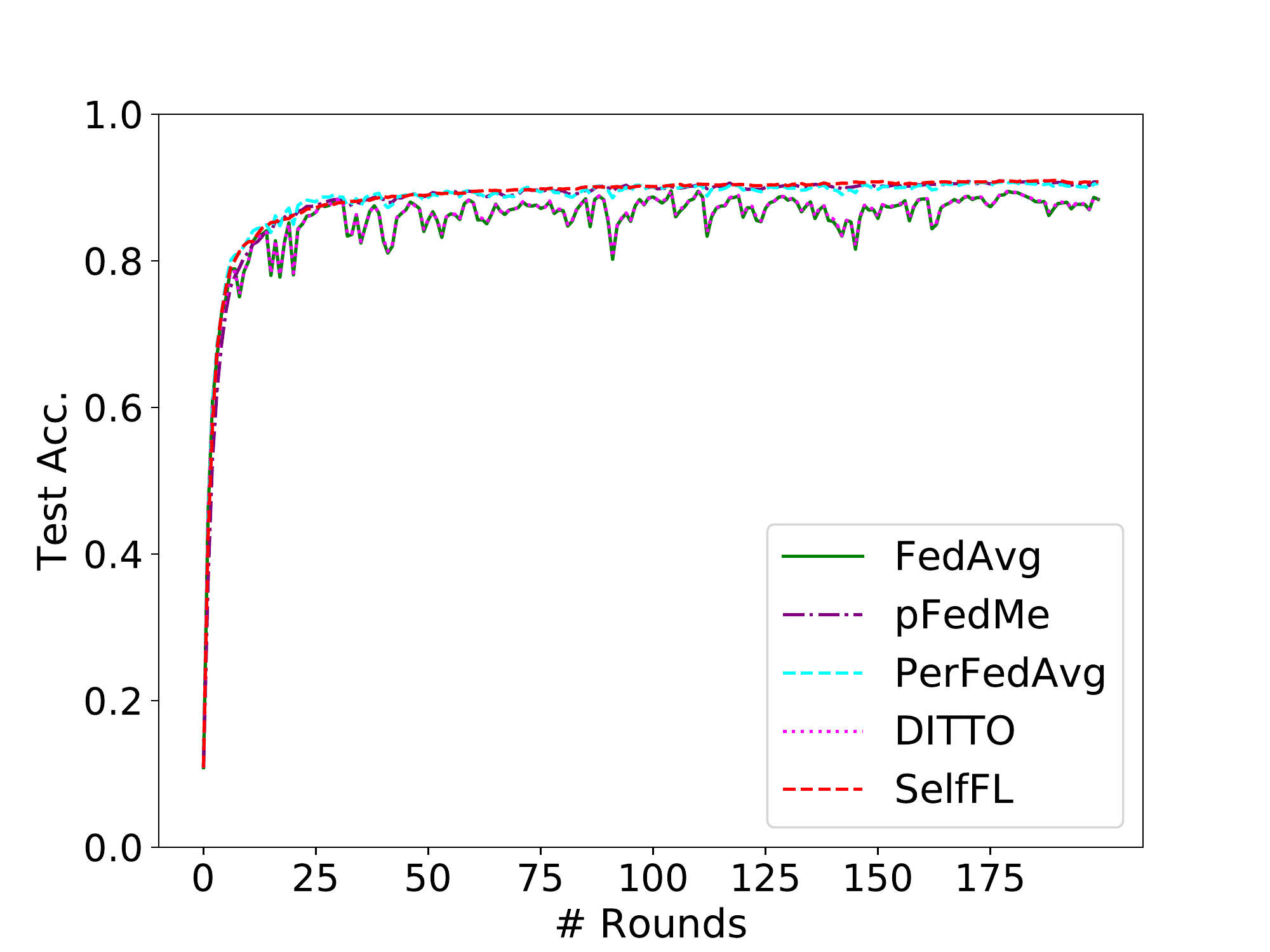}  
  \caption{MNIST}
  \label{fig:comp_mnist}
\end{subfigure}
\begin{subfigure}[b]{0.49\columnwidth}
  \centering
  \includegraphics[width=\linewidth]{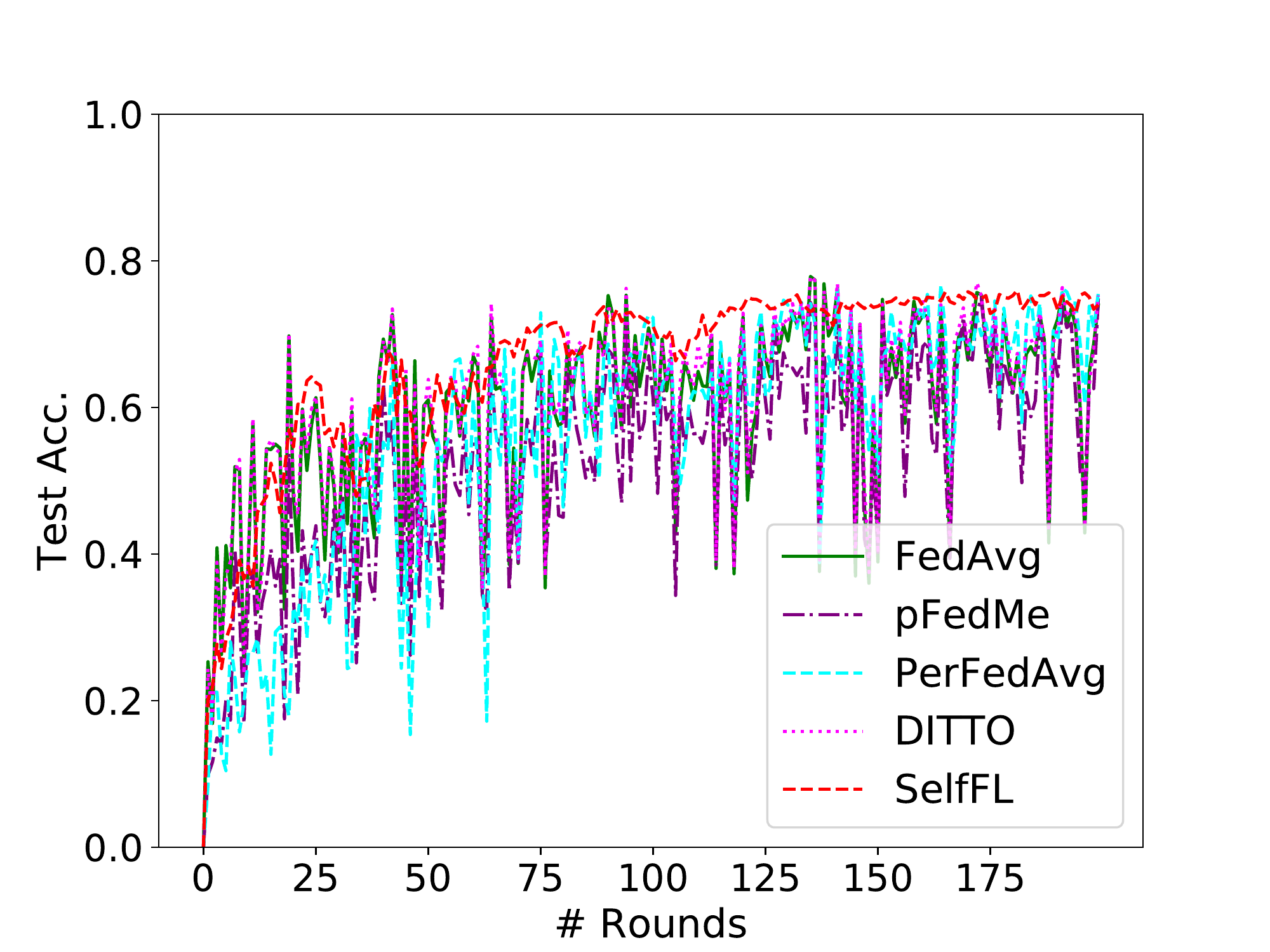}  
  \caption{FEMNIST}
  \label{fig:comp_femnist}
\end{subfigure}
\vspace{-1.1em}
\caption{The weighted client-level test accuracy in different communication rounds (with $0.1$ activity rate). }
\label{fig:comp_images}
\end{figure}

\vspace{-0.2em}
Figures~\ref{fig:comp_mnist} and~\ref{fig:comp_femnist} compare the convergence performance of different FL algorithms on MNIST and FEMNIST, respectively. The horizontal axis denotes the communication round. 
The vertical axis denotes the weighted test
accuracy of individual clients, where the weight is proportional to the sample size.  
We observe that: 
(i) \sys{} algorithm yields more \textit{stable convergence} compared with FedAvg and the other three personalized FL methods due to our smoothing average in aggregation; 
(ii) \sys{} achieves the highest personalized accuracy when the global model converges.

To corroborate that the performance advantage of \sys{} does not come from a larger potential local training steps (we set $l_{max}=40$ for \sys{}), we perform an additional experiment where we use a local step as $l=40$ (instead of $20$) for DITTO, pFedMe, and PerFedAvg on the FEMNIST dataset. In this scenario, the test accuracy of these three FL algorithms is $71.33\%$, $68.87\%$, and $71.08\%$, respectively. Our proposed method achieves a test accuracy of $76.93\%$. Furthermore, it is worth noting that other FL methods do not have a principled way to determine the value of $l$. The adaptive local steps in \sys{} also address this challenge.



\begin{figure}[ht!]
\begin{subfigure}[b]{0.49\columnwidth}
  \centering
  \includegraphics[width=\linewidth]{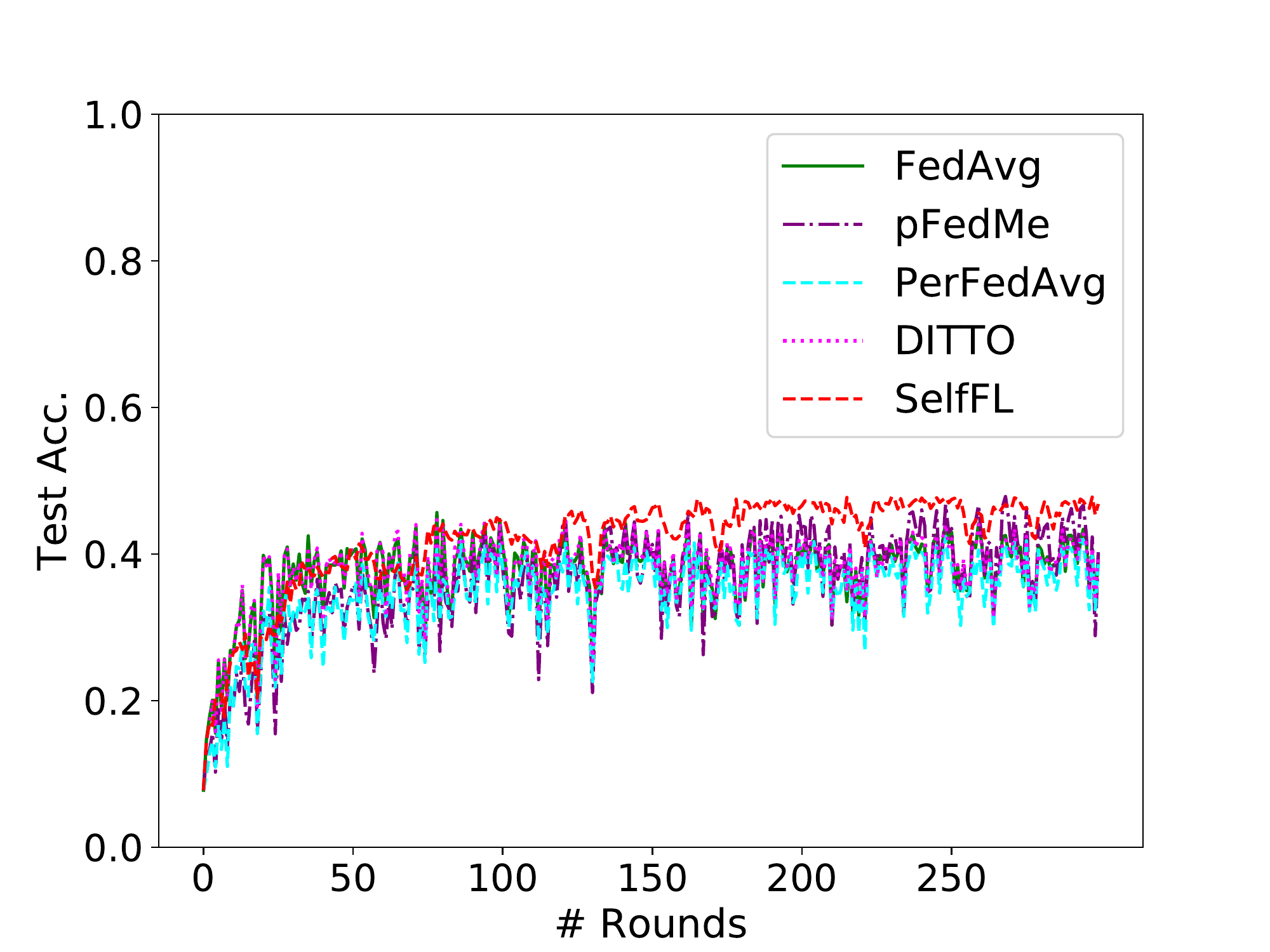}  
  \caption{CIFAR10}
  \label{fig:comp_cifar10}
\end{subfigure}
\begin{subfigure}[b]{0.49\columnwidth}
  \centering
  \includegraphics[width=\linewidth]{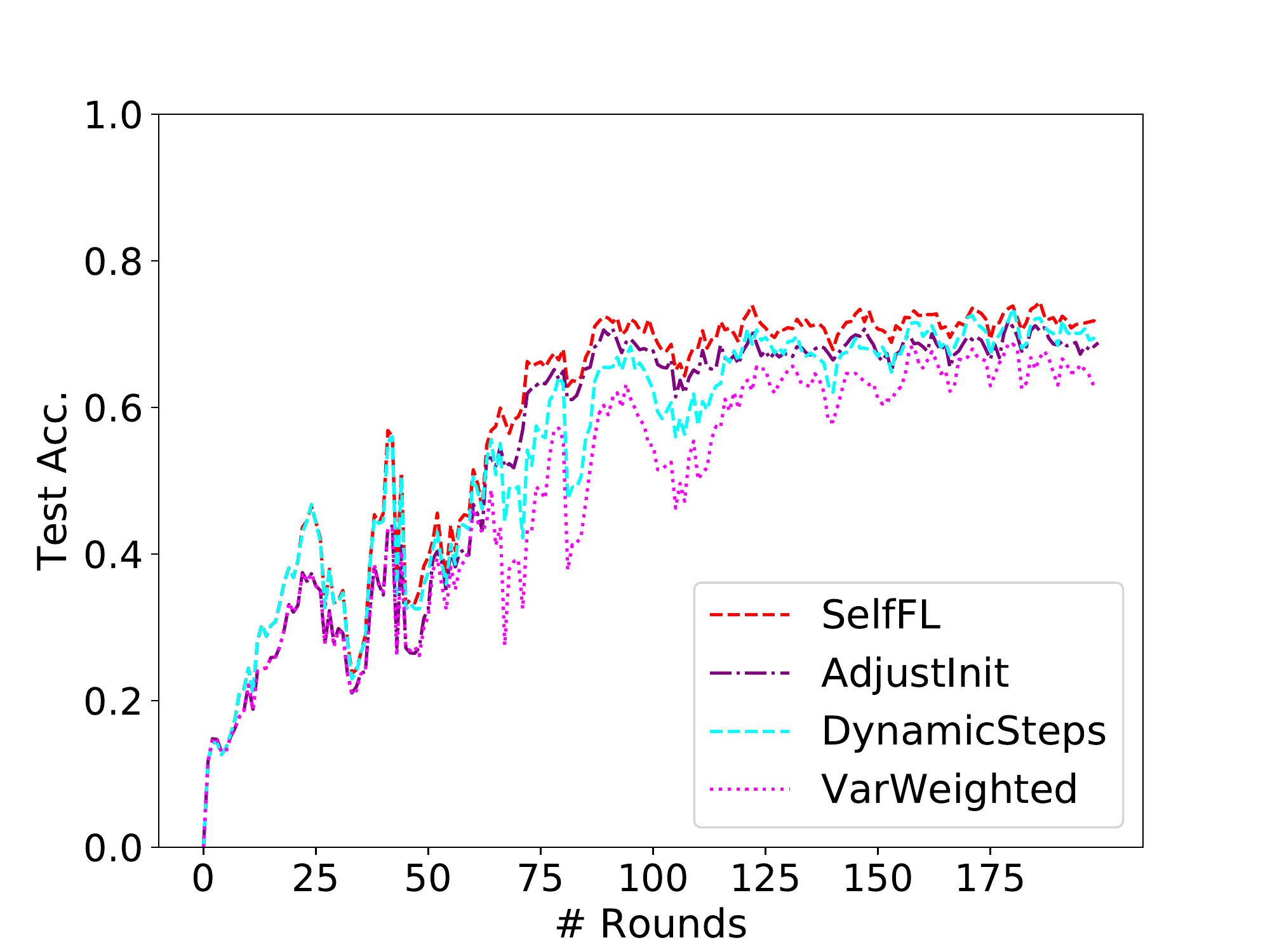}  
  \caption{FEMNIST}
  \label{fig:ablation_sys_femnist}
\end{subfigure}
\vspace{-0.8em}
\caption{Weighted client-level test accuracy. }
\label{fig:comp_cf10_nist}
\end{figure}

\noindent \textbf{Results on CIFAR10.} We also perform experiments on the CIFAR-10 dataset (from the pFedMe repository~\cite{pfedme_repo}) with the CNN model from PyTorch example~\cite{pytorch_cf10}. In each round, we randomly select 5 out of 20 clients to participate in FL training. We use a learning rate of $0.01$ for all FL algorithms and set the local training step as $40$ for the baseline methods. Figure~\ref{fig:comp_cifar10} shows the test accuracy comparison between different FL methods, showing that \sys{} achieves a higher accuracy.

\vspace{0.2em}
\noindent \textbf{Ablation study on \sys{}.} Recall that \sys{} consists of three key components: (i) Adjusting local initialization; (ii) Dynamic local training steps; (iii) Variance-weighted global aggregation (Section~\ref{sec_method}). To investigate the effect of each component, we perform an ablation study where we apply the full deployment of \sys{} and only only component of it.
Figure~\ref{fig:ablation_sys_femnist} compares the test accuracy of the global model on FEMNIST benchmark in these four settings. We can observe that adjusting the local initialization at each round (Eqn.~(\ref{eq_101})) gives the highest contribution to \sys{}'s performance.

To characterize the distribution of personalized performance across clients, we define two new metrics to evaluate different FL algorithms: (i) weighted test accuracy of clients that have top 10\% most samples; (ii) worst 10\% clients' average test accuracy. The evaluation results on FEMNIST (where the client sample size ranges from 1 to 206) are shown in Table~\ref{tab:new_metrics} (standard errors in parentheses). We can observe that \sys{} outperforms other FL algorithms in terms of these additional metrics.

\begin{table}[ht!]
\centering
\caption{Test accuracy comparison on FEMNIST. \label{tab:new_metrics}}
\vspace{0.3em}
\scalebox{0.75}{
\begin{tabular}{c|c|c|c|c|c} 
\toprule
\textbf{FL Alg.} & \textbf{FedAvg} & \textbf{DITTO} & \textbf{pFedMe} & \textbf{PerFedAvg} & \textbf{Self-FL} \\ 
\midrule
\begin{tabular}[c]{@{}c@{}}Top 10\% \\most samples\end{tabular} & \begin{tabular}[c]{@{}c@{}}70.48\% \\(0.07\%)\end{tabular} & \begin{tabular}[c]{@{}c@{}}70.58\% \\(0.14\%)\end{tabular} & \begin{tabular}[c]{@{}c@{}}72.76\%\\(0.28\%)\end{tabular} & \begin{tabular}[c]{@{}c@{}}74.95\%\\(0.56\%)\end{tabular} & \begin{tabular}[c]{@{}c@{}}76.24\%\\(0.07\%)\end{tabular} \\ 
\hline
\begin{tabular}[c]{@{}c@{}}Worst 10\% \\user avg acc\end{tabular} & \begin{tabular}[c]{@{}c@{}}0\% \\(0\%)\end{tabular} & \begin{tabular}[c]{@{}c@{}}0\%\\(0\%)\end{tabular} & \begin{tabular}[c]{@{}c@{}}17.97\%\\(1.23\%)\end{tabular} & \begin{tabular}[c]{@{}c@{}}24.51\%\\(1.55\%)\end{tabular} & \begin{tabular}[c]{@{}c@{}}37.01\%\\(1.57\%)\end{tabular} \\
\bottomrule
\end{tabular}
}
\vspace{-0.6em}
\end{table}

\vspace{-0.4em}
\subsection{Effectiveness on the audio domain} \label{sec:eval_audio}
\vspace{-0.3em}
In this section, we evaluate the performance of \sys{} on an audio dataset (Subsection~\ref{sec:setup}) for wake-word detection. 
Particularly, we use a CNN that is pre-trained on the training data of different device types (i.e., heterogeneous data) as the initial global model to \textit{warm-start} FL training for all evaluated FL algorithms. The personalization task aims to improve the wake-word detection performance at the device type level. 
The output value from the CNN is compared with a pre-defined threshold to determine if the wake-word is present in the input audio. 
We implement \sys{} based on Algorithm~\ref{algo} to obtain the global and local models. 

\vspace{-0.4em}
The smart devices have three operation states that impact the wake-word detection performance: normal, playback, and alarm. The playback and alarm states respectively mean that the device is playing music and alarming during the detection. 
The sample sizes are summarized in Table~\ref{tab:ww_stats}. We target five device types denoted by `A'$\sim$`E' in this experiment. 
We use a batch size of $500$ for all clients.

\begin{table}[ht!]
\centering
\caption{Sample sizes in the wake-word detection dataset. \label{tab:ww_stats}}
\vspace{0.3em}
\scalebox{0.78}{
\begin{tabular}{ccccccc} 
\toprule
\multicolumn{2}{c}{\multirow{2}{*}{\textbf{\# Audio Streams}}} & \multicolumn{5}{c}{\textbf{Device Type}} \\
\multicolumn{2}{c}{} & \textbf{A} & \textbf{B} & \textbf{C} & \textbf{D} & \textbf{E} \\ 
\midrule
\multirow{3}{*}{\textbf{State}} & \textbf{normal} & 43922 & 28852 & 10404 & 76804 & 51097 \\
 & \textbf{playback} & 12011 & 4223 & 4336 & 21511 & 10985 \\
 & \textbf{alarm} & NA & 646 & NA & 1427 & 1318 \\
\bottomrule
\end{tabular}
}
\end{table}

\noindent \textbf{Evaluation metric.}
One can control the trade-off between false accepts and false rejects of wake-word detection by tuning the threshold. Since we use a pre-trained model to warm-start FL, we first evaluate the detection performance of this pre-trained model as the baseline. 
To compare the performance of different FL algorithms, we use the \textit{relative false accept (FA)} value of the resulting model when the corresponding relative false reject (FR) is close to one as the metric. So a relative FA smaller than one is preferred. Here, the relative FA and FR are computed with respect to the baseline. We detail the results in two scenarios below. 

In this experiment, we assume there are five clients in the FL system. Each client only has the training data for a specific device. There is no data overlap between clients. Furthermore, all the clients participate in each communication round. 
Note that FedAvg~\citep{mcmahan2017communication} and DITTO~\citep{li2021ditto} typically use the sample size-based weighted average to aggregate models, which do not take data heterogeneity into account. 
For comparison, we implement FedAvg and DITTO with both equal-weighted model averaging (denoted by the suffix `-e') and sample size-based weighted averaging (denoted by the suffix `-w') during aggregation. For the meta-learning-based method PerFedAvg~\citep{perfedavg}, we use its first-order approximation and the equal-weighted aggregation. 
{We did not report pFedMe on this dataset because we found it did not converge with various hyper-parameters. }

\vspace{-0.2em}
We evaluate the performance when devices are in a normal state. Table~\ref{tab:ww_compare_full} summarizes the performance of the updated global model.
Recall that a smaller relative FA indicates better performance. Each column reports the relative FA for a specific device type. 
The results show that \sys{} achieves the lowest relative FA. 
{Several updated global models have worse performance than the baseline model, e.g., for {FedAvg-w} and {DITTO-w} in Table~\ref{tab:ww_compare_full}. This is due to the setting here that local models are initialized from a pre-trained model, and the comparison is relative to that warm-start. Because clients' data are highly heterogeneous, a sample size-based aggregation rule used in FedAvg-w and Ditto-w results in a deteriorated global model compared with the original start.}
We provide ablation studies in the client sampling setting in the Appendix Subsection~\ref{app:ww_sampling}, which show \sys{} still performs well.

\begin{table}[tb]
\centering
\caption{Detection performance (relative FA) of the \textit{global model} 
on a test dataset (around 20\% the size of training data), when the devices are in the normal state.  \label{tab:ww_compare_full} }
\vspace{0.4em}
\scalebox{0.78}{
\begin{tabular}{cccccc} 
\toprule
\multirow{2}{*}{\textbf{FL methods}} & \multicolumn{5}{c}{\textbf{Device Types}} \\
 & \textbf{A} & \textbf{B} & \textbf{C} & \textbf{D} & \textbf{E} \\ 
\midrule
\textbf{Self-FL} & \multicolumn{1}{l}{\textbf{0.92}} & \multicolumn{1}{l}{\textbf{0.94}} & \multicolumn{1}{l}{\textbf{0.91}} & \multicolumn{1}{l}{\textbf{0.91}} & \multicolumn{1}{l}{1.01} \\
\textbf{FedAvg-w} & 8.39 & 4.00 & 12.80 & 8.61 & 10.62 \\
\textbf{FedAvg-e} & 0.97 & 0.96 & 1.00 & 0.92 & 1.00 \\
\textbf{DITTO-w} & 8.38 & 4.00 & 12.75 & 8.61 & 10.23 \\
\textbf{DITTO-e} & 0.97 & 0.95 & 1.00 & 0.93 & \textbf{0.99} \\
\textbf{PerFedAvg} & 1.06 & 0.98 & 1.08 & 0.93 & 1.01 \\
\bottomrule
\end{tabular}
}
\end{table}

\vspace{-0.2em}
We further compare the personalization performance of our \sys{} framework with FedAvg-e and two personalized FL algorithms, DITTO-e and PerFedAvg, on each device type's test data and summarize the results in Table~\ref{tab:full_local_model}. We do not consider FedAvg-w and DITTO-w in this experiment since they result in performance degradation, as seen from Table~\ref{tab:ww_compare_full}.
One can see that \sys{} outperforms the other methods across all device types, thus demonstrating a better personalization capability.

\vspace{-1.2em}
\begin{table}[ht!]
\centering 
\caption{
Detection performance (relative FA) of the \textit{personalized models} on a test dataset, when the devices are in the normal state. \label{tab:full_local_model} }
\vspace{-0.2cm}
\scalebox{0.78}{
\begin{tabular}{cccccc} 
\toprule
\multirow{2}{*}{\textbf{FL methods}} & \multicolumn{5}{c}{\textbf{Device Type}} \\
 & \textbf{A} & \textbf{B} & \textbf{C} & \textbf{D} & \textbf{E} \\ 
\midrule
\textbf{\sys{}} & \textbf{0.93} & \textbf{0.91} & \textbf{0.90} & \textbf{0.90} & 0.99 \\
\textbf{FedAvg-e} & 0.95 & 0.95 & 0.93 & 0.91 & 0.98 \\
\textbf{DITTO-e} & 0.97 & 0.96 & 0.93 & 0.91 & 0.96 \\
\textbf{PerFedAvg} & 1.02 & 1.11 & 1.08 & 1.00 & \textbf{0.93} \\
\bottomrule
\end{tabular}
}
\end{table}

\vspace{-0.8em}
\section{Concluding Remarks}
\vspace{-0.5em}
Personalized FL has many potential applications. In this paper, we proposed \sys{} to address the challenge of balancing local model regularization and global model aggregation. Its key component is using uncertainty-driven local training steps and aggregation rules instead of conventional local fine-tuning and size-based weights. To our best knowledge, \sys{} is the first work that establishes the connection between personalized FL and hierarchical modeling and utilizes uncertainty quantification to drive the personalization. Extensive empirical evaluations of \sys{} show its promising performance. 
%
%

There are some new problems left from the work that deserve further research. 
First, in many practical applications, clients only have unlabeled data, and it is not easy to annotate those data. This poses a significant challenge in training personalized models and evaluating them. An important problem is to integrate self-supervised FL techniques~\cite{SemiFL} into personalization to address such a lack-of-label issue.
Second, we considered heterogeneity in terms of clients' data distributions. There are other aspects of heterogeneity, e.g., those at the system and model levels. A more integrated view of FL heterogeneity is lacking. Third, a critical problem in personalized FL is to assess whether any particular client model, for a given set of local data, has achieved its theoretical limit from, e.g., a goodness-of-fit perspective~\cite{DingBAGofT}.
We refer to \cite{ding2022federated} for an outlook on other challenges related to personalized FL.

\vspace{-0.3em}
The \textbf{Appendix} contains further experimental studies, remarks, and technical proofs.

\balance
\bibliography{privacy,J}
\bibliographystyle{icml2022}

\newpage
\onecolumn
\appendix
\pagenumbering{gobble}

\centerline{\LARGE Appendix for ``Self-Aware Personalized Federated Learning''} 

\vspace{1cm}

This \textbf{Appendix} is structured as follows. We provide a summary of frequently used notations in Section~\ref{app:notations} and additional related works in Section~\ref{app:related}.
In Section~\ref{app:implemen_details}, we discuss how the uncertainty quantity in Algorithm~\ref{algo} can be computed in an online fashion.
In Section~\ref{app:complex_alg}, we introduce a variant of \sys{} named ``Augmented \sys{}'', which serves as a natural alternative to Algorithm~\ref{algo}. 
Then, in Section~\ref{app:addition_results}, we conduct ablation studies and provide additional experimental results. Finally, we provide the proof of Proposition~\ref{prop_converge} in Section~\ref{app:proof_proposition}.

\vspace{-0.6em}
\section{Summary of Notations}  \label{app:notations}
\vspace{-0.3em}
We summarize the frequently used notations in Table~\ref{tab_notation}.

\vspace{-1.2em}
\begin{table*}[!th]
\caption{Frequently used notations} \label{tab_notation}
\vspace{0.5em}
\begin{tabular}{lll}
\hline
Notation            & Meaning                                                                                    &  \\ \hline
$M$                 & number of clients                                                                          &  \\
$\vD_m$             & variance of client $m$'s estimated parameters, or the ``intra-client uncertainty'' in general                                              &  \\
$\vT$               & variance of different clients' underlying parameters, or the ``inter-client uncertainty'' in general                                                 &  \\
$\thL_m$, $\vL_m$   & posterior mean and variance of client $m$'s personal parameter                             &  \\
$\thFL_m$, $\vFL_m$ & posterior mean and variance of client $m$'s personal parameter conditional on all the data &  \\
$\thG$, $\vG$       & global model's posterior mean and variance                                                 &  \\
$\eta$              & learning rate                                                                        &  \\
$\ell$              & learning steps                                                                       &  \\
$t$                 & FL round                                                                                   &  \\
$\th_{m}^{t}$       & client $m$'s FL-optimal parameter                                                          &  \\
$\th_{m, t}$        & client $m$'s local optimal parameter at round $t$                                                      &  \\ \cline{3-3} 
$\th^t$             & global parameter at around $t$                                                                  &  \\ \hline
\end{tabular}
\end{table*}

\section{Further Remarks on Related Work}  \label{app:related}

\textbf{Federated learning.} A general goal of FL is to train massively distributed models at a large scale~\citep{bonawitz2019towards}. FedAvg~\citep{mcmahan2017communication} is perhaps the most widely adopted FL baseline, which reduces communication costs by allowing clients to train the local models for multiple iterations. To further reduce communication costs, data compression techniques such as quantization and sketching have been developed for FL~\citep{konevcny2016federated,alistarh2017qsgd,ivkin2019communication}. To further reduce computation costs, techniques to train a large model using small-capacity devices such as HeteroFL~\citep{DingHeteroFL} have been developed.

\textbf{Bayesian Federated Learning.} 
There have been prior work that use the Bayesian perspective in FL. For instance, FedPA~\cite{al2020federated} formulates FL as a posterior inference problem where the global posterior is obtained by averaging the clients' local posteriors. FedBE~\cite{chen2020fedbe} proposes an aggregation method by interpreting local models as samples from the global model distribution and leveraging Bayesian model ensemble to aggregate the predictions.
There are two main differences between \sys{} and the above works. First, both FedPA and FedBE aim to learn a global FL model on the server side instead of personalized models for clients (the focus of our paper). Thus, the developed update rules and use scenarios are entirely different. Second, our approach is based on a novel two-level hierarchical Bayes perspective that leverages the inter-client and intra-client uncertainty to drive FL optimization. FedPA and FedBE use standard (one-level) Bayesian posterior to update the global model. 

\textbf{Adaptive Federated Learning.} 
There has been a direction of research to automate the hyperparameter tuning process in FL. For example, FedOPT~\cite{reddi2020adaptive} proposes an FL framework with server and client optimizers to improve FL convergence instead of personalization (our focus). The adaptive server optimization in~\cite{reddi2020adaptive} is orthogonal to Self-FL. FedOPT involves tuning hyperparameters for the server’s adaptive optimizer (e.g., selecting $\beta_1$, $\beta_2$, and $\epsilon$ for Adam). Also, its aggregation rule is the same as standard FL and can be possibly replaced by our method for better performance. 
FedEx~\cite{khodak2021federated} proposes a hyperparameter tuning method inspired by weight sharing in neural architecture search. Although FedEx can auto-tune local hyperparameters, it may need extra efforts to determine its own hyperparameters, such as $\eta_t$, $\lambda_t$, and configurations $c$. 
In terms of the update rules, FedEx and Self-FL are also very different. In particular, a client in FedEx will sample a configuration $c$ from a hyperparameter distribution $D$, and the server will perform an exponentiated update for $D$; Self-FL will estimate empirical variances (online) to simultaneously determine local initialization, training steps, and global model aggregation.

\vspace{-0.7em}
\section{Implementation Details of Algorithm~\ref{algo}} \label{app:implemen_details}
\vspace{-0.3em}

In Algorithm~\ref{algo}, we need to evaluate the empirical variance of a set of models. This can be a memory concern for edge devices in practice. We use the following way to perform online calculation, so that the required hardware memory does not grow with the number of rounds or clients.

To calculate the sample variance 
\begin{align}
    \widehat{\sigma^2_t} \de \frac{1}{t}\sum_{i \in [t]} (x_i-\bar{x}_t)^2 \label{eq19}
\end{align}
where $\bar{x}_t \de t^{-1}\sum_{i \in [t]} x_i$,
we do not need to store $x_1,\ldots,x_t$. Instead, we only need to store $\bar{x}_{t}$ at time step $t$, and calculate the sample variance in the following recursive way.
\begin{align}
  &\textrm{For }t=1,2,\ldots \nonumber \\
  &\bar{x}_{t} = \frac{t-1}{t} \bar{x}_{t-1} + \frac{1}{t} x_t, \nonumber \\
  &\widehat{\sigma^2_t} = \frac{t-1}{t} \widehat{\sigma^2_{t-1}} + \frac{t-1}{t} (\bar{x}_{t} - \bar{x}_{t-1}) + \frac{(x_t-\bar{x}_{t})^2}{t} ,\label{eq20}
\end{align}
with $\bar{x}_{0}=\widehat{\sigma^2_0}=0$. 
It can be verified that the $\widehat{\sigma^2_t}$ in (\ref{eq20}) is equivalent to that in (\ref{eq19}).
The recursive computation is particularly favorable for large neural networks with millions of parameters and small hardware memory. 

\begin{remark}[Client sampling]
    \label{remark_client_sampling}
    Suppose that in each round, only a subset of clients is activated. The subscript $t$ in the online update of $\sigma^2_m$ is client $m$'s the local time counter, namely the total counts that client $m$ is activated in FL communications. This local time counter shall be distinguished from the global counter $t$ that indicates the system communication round. Also, a client's intra-client uncertainty is only online updated in those rounds that it is activated.

\end{remark}


\section{An Alternative Version of Algorithm~\ref{algo}} \label{app:complex_alg}

In this section, we present a slightly more complicated version of \sys{} in Algorithm~\ref{algo_ablation_complex_version}, as an alternative of Algorithm~\ref{algo}. Its pseudocode can be found at the end of the Appendix, and the main changes are highlighted with blue fonts.
The motivation goes back to the discussions in Subsection~\ref{subsec_algo}, where we mentioned the use of a client $m$'s local parameter at each round as a bootstrapped estimate of the underlying $\thL_m$.
In Algorithm~\ref{algo_ablation_complex_version}, each client needs to run additional local training until convergence (denoted by $\th_{m,t}$) in each FL round and communicates two local models with the server. 

This alternative variant requires the computation and communication of local-optimal solutions ($\th_{m,t}$). Thus, the implementation of \sys{} in Algorithm~\ref{algo_ablation_complex_version} incurs higher computation costs for local clients (due to additional local fine-tuning) and doubles the communication overhead compared with the standard FedAvg~\citep{mcmahan2017communication}. 
We will perform ablation studies in Subsection~\ref{app:ablation} to compare these two algorithms.

\begin{algorithm*}[!ht]
\SetAlgoLined
\DontPrintSemicolon
\caption{Self-Aware Personal Federated Learning by Introducing Local Minimum ${\color{blue} \th_{m,t}}$ (Augmented Self-FL). \\ The main differences with Algorithm~\ref{algo} are highlighted with {\color{blue} blue} fonts.}
\label{algo_ablation_complex_version}
\Input{A server and $M$ clients. System-wide objects, including the initial model parameters $\th_{m}^{0} = \th_{m, 0} = \th^0$, initial uncertainty of clients' parameters $\widehat{\vT}=0$, the activity rate $C$, the number of communication rounds $T$, server batch size $B_s$, and client batch size $B_m$, and the loss function $(z, \th) \mapsto \loss(z, \th)$ corresponding to a common model architecture. Each client $m$'s local resources, including $N_m$ data observations, parameter $\th_m$, number of local training steps $\ell_m$, and learning rate $\eta_m$. 
}
\kwSystem{}{
    \For{\textup{each communication round }$t = 1,\dots T$}{

        $\mathbb{M}_t \leftarrow \max(\lfloor C \cdot M \rfloor ,1)$ active clients uniformly sampled without replacement
    
        \For{\textup{each client }$m \in \mathbb{M}_t$\textup{ \textbf{in parallel}}}{
    
            Distribute server model parameters $\th^{t-1}$ to local client $m$
    
            $(\th_{m}^{t}, {\color{blue} \th_{m,t}}, \widehat{\vD_m}) \leftarrow$ \textbf{ClientUpdate}$(z_{m,1},\ldots,z_{m,N_m}, \th^{t-1}, \widehat{\vT})$
        }
        
        $(\th^t, \widehat{\vT}) \leftarrow$ \textbf{ServerUpdate}$(\th_{m}^{t}, {\color{blue} \th_{m,t}}, \widehat{\vD_m}, m \in \mathbb{M}_t)$
    }
}
\kwClient{\textup{$(z_{m,1},\ldots,z_{m,N_m}, \th^{t-1}, \widehat{\vT})$}}{

    Calculate the uncertainty of the local optimal solution $\thL_m$ of client $m$:
    \begin{align}
     \widehat{\vD_m} = \textrm{empirical variance of }\{{\color{blue} \th_{m,1}, \ldots , \th_{m,t-1} } \} .
    \end{align}
    
    Calculate the initial parameter 
    \begin{align}
    \th_{m}^{t} &\de \th^{t-1} -  \frac{(\widehat{\vT} + \widehat{\vD_m})^{-1} }{\sum_{k: k \neq m} (\widehat{\vT} + \widehat{\vD_k})^{-1}} ({\color{blue} \th_{m, t-1}} - \th^{t-1}) ,
    \end{align}

    Choose $\ell_m \geq 1$ according to 
    \begin{align}
    \biggl(1-\frac{\eta_m }{B_m \widehat{\vD_m}} \biggr)^{\ell_m}
    &=  \frac{ \sum_{k: k \neq m} (\widehat{\vT} + \widehat{\vD_k})^{-1}}{ 1/\widehat{\vD_m} + \sum_{k: k \neq m} (\widehat{\vT} + \widehat{\vD_k})^{-1} } \nonumber
    \end{align}

    \For{\textup{local step $s$ from $1$ to $\ell_m$}}{
    
        Sample a batch $I \subset [N_m]$ of size $B_m$
        
        Update $\th_{m}^{t} \leftarrow \th_{m}^{t} - \eta_m \nabla_{\th} \sum_{i \in I} \loss (\w_{m,i}, \th_{m}^{t}), $
    }

    Let ${\color{blue} \th_{m,t}} \leftarrow \th_{m}^{t}$.
    
    {\color{blue}
    \For{\textup{local step $s$ from $\ell_m+1$ to $\infty$}}{
    
        Sample a batch $I_s \subset [N_m]$ of size $B_m$
        
        Update $\th_{m,t} \leftarrow \th_{m,t} - \eta_m \nabla_{\th} \sum_{i \in I_s} \loss (\w_{m,i}, \th_{m,t}), $
        
        If {it converges}, then Break.
    }

    Store ${\color{blue} \th_{m,t}}$ locally
    
    } 
    
    Return $(\th_{m}^{t}, {\color{blue} \th_{m,t}}, \widehat{\vD_m})$ and send them to the server
}
\kwServer{\textup{$(\th_{m}^{t}, {\color{blue} \th_{m,t}}, \widehat{\vD_m}, m \in \mathbb{M}_t)$}}{
    Calculate the uncertainty of clients' underlying parameters
    \begin{align}
        \widehat{\vT} = \textrm{empirical variance of }\{{\color{blue} \th_{m,t}}, m \in \mathbb{M}_t\} .
    \end{align}
        
    Calculate model parameters
        \begin{align}
            \th^t &\de \frac{\sum_{m \in \mathbb{M}_t} (\widehat{\vT} + \widehat{\vD_m})^{-1} {\color{blue} \th_{m,t}}}{ \sum_{m \in \mathbb{M}_t} (\widehat{\vT} + \widehat{\vD_m})^{-1} } .
        \end{align}
    
    Return $(\th^t, \widehat{\vT})$
}
\end{algorithm*}

\vspace{-0.5em}
\section{Additional Experiments}  \label{app:addition_results}
\vspace{-0.5em}
We perform extended experiments of \sys{} and summarize the results in this section. Particularly, in Subsection~\ref{app:ablation}, we compare the performance of two implementation variants of \sys{}. In Subsection~\ref{app:ww_sampling}, we revisit the audio data in a client sampling setting, as a supplement to the full participation setting studied in Subsection~\ref{sec:eval_audio}. 
In Subsections~\ref{app:activity_rate} and~\ref{app:eval_theoretical_lm}, we show the impact of the client activity rate $C \in (0,1]$ and provide insights into the adaptive local training steps of \sys{}, respectively.

\vspace{-0.5em}
\subsection{Convergence study on synthetic data} \label{sec:eval_synthetic}
\vspace{-0.5em}

{For FedAvg on the synthetic dataset, we use a gradient descent-based local update rule: $\theta^l_1=\theta^{l-1}_1-\eta(\theta^{l-1}_1 - \theta^{(L)}_1)/\sigma^2_1$. For DITTO, an additional regularization term is used and the update rule is: $\theta^l_1=\theta^{l-1}_1-\eta(\theta^{l-1}_1 - \theta^{(L)}_1)/\sigma^2_1 -\lambda(\theta^{l-1}_1-\theta^{(G)})$. 
A skeptical reader may wonder whether DITTO is equivalent to Self-FL for a particular choice of $\lambda$. This is not the case, even in our earlier example of the two-layer Gaussian model. The two methods differ in the aggregation rule and the initialization of each client. 
}



\noindent \textbf{Case 1: Homogeneous setting. }
We use a small inter-client uncertainty in this setup and assume that clients have similar local sample sizes. There are $M=20$ clients in the FL system where the local sample size $N_m$ is uniformly sampled from the range $[10, 20]$. We set the parent parameter $\theta_0=1.6$. The inter-client and intra-client uncertainty are $\sigma^2_{0}=0.001$,  $\sigma^2_{m}=0.1$, respectively.
The experiment is repeated $200$ times, and we report the mean value as the final metric. The standard error of the experiment is within $0.001$. Note that the effective $\sigma^2_{m}$ shall be divided by $N_m$.

\vspace{1em}
\begin{figure}[b]
\begin{subfigure}[b]{0.49\columnwidth}
  \centering
  \includegraphics[width=\linewidth]{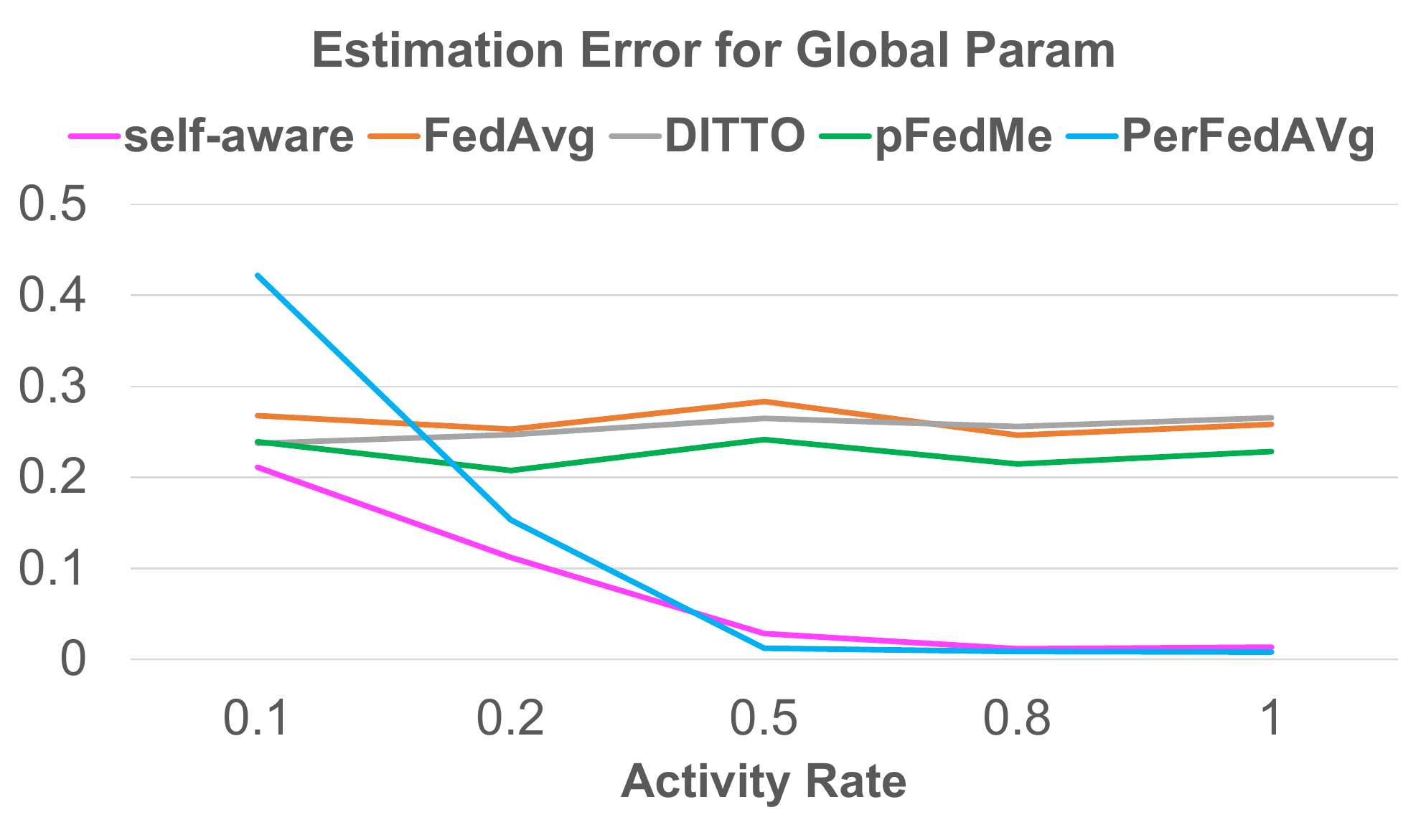}  
  \caption{}
  \label{fig:homo_global}
\end{subfigure}
\hfill
\begin{subfigure}[b]{0.49\columnwidth}
  \centering
  \includegraphics[width=\linewidth]{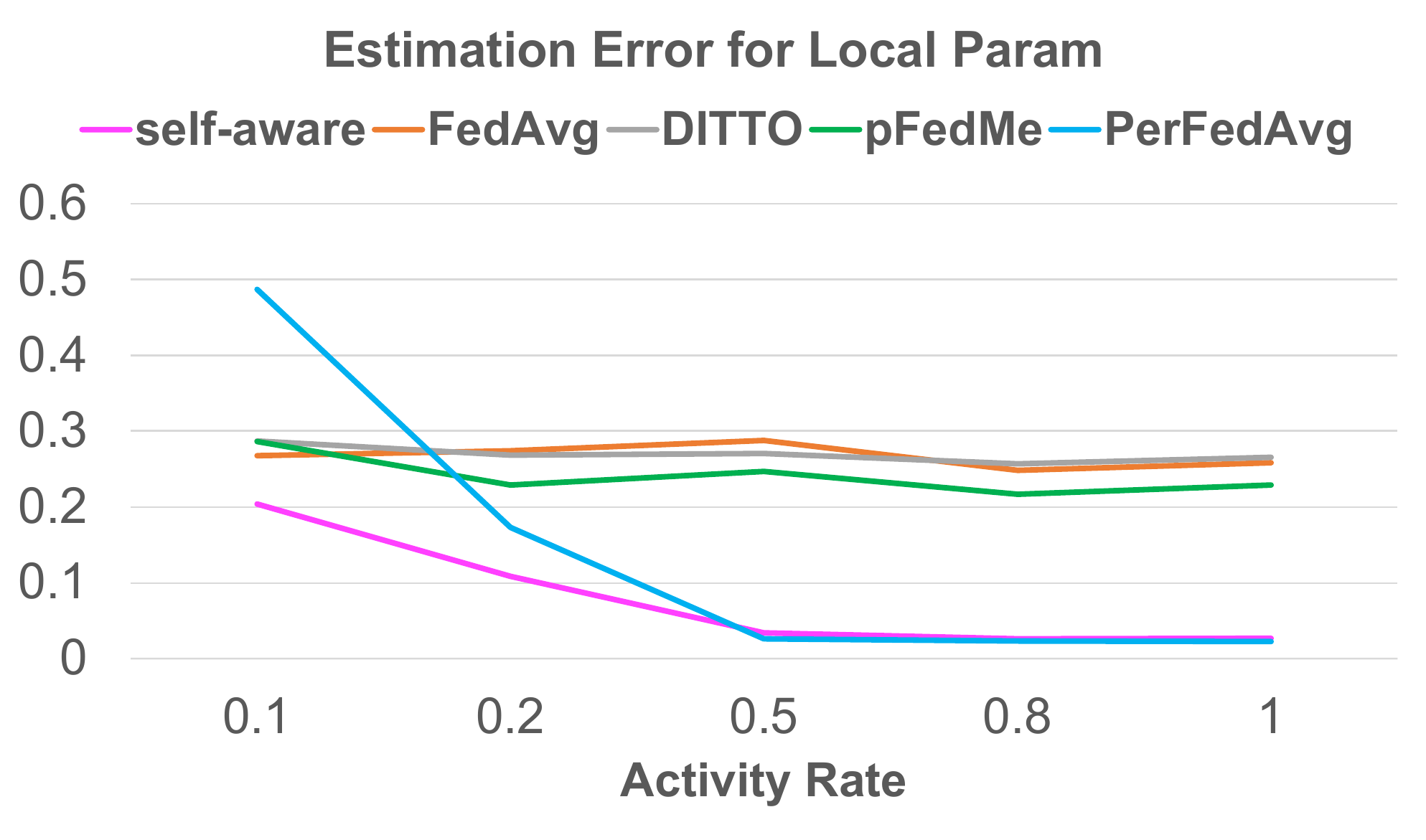}  
  \caption{}
  \label{fig:homo_local}
\end{subfigure}
\vspace{-1.2em}
\caption{Performance of different FL algorithms on synthetic data of high homogeneity. Estimations errors of the global and local parameters (averaged over clients) are shown in (a) and (b), respectively.  
}
\label{fig:comparison_homo}
\end{figure}

\vspace{-1em}
Figure~\ref{fig:comparison_homo} compares the parameter estimation performance of \sys{} algorithm with FedAvg and three existing personalized FL techniques: DITTO~\citep{li2021ditto}, pFedMe~\citep{pfedme}, and PerFedAvg~\citep{perfedavg}. The estimation error is measured in $\ell_1$ norm. 
All FL algorithms use the same learning rate $10^{-4}$ and are trained for $200$ rounds.
For \sys{}, the local training steps $l_m$'s are computed using Eqn.~(\ref{eq:lm}). For other FL methods, the local training steps are set to $50$.
We consider the activity rate $C \in \left\{0.1, 0.2, 0.5, 0.8, 1 \right\}$ and compare different algorithms in each setting. 
We draw two conclusions from Figure~\ref{fig:comparison_homo}. (i) When there are sufficient active clients in each communication round, \sys{} and PerFedAvg achieve the lowest estimation error for both global and local parameters compared with others. (ii) When the number of active clients is small, \sys{} outperforms the other methods. 

\noindent \textbf{Case 2: Heterogeneous setting. }
To generate heterogeneous data across clients, we use a large value for inter-client uncertainty $\sigma^2_{0}$ and assume that clients have different scales of local samples size $N_m$. We consider $\sigma^2_{0}=1$ and $\sigma^2_{m}=0.1$. 
The sample size of each client is uniformly drawn from $[10, 200]$ (which is much broader than that in Case 1). 
Other settings are the same as Case 1.

\begin{figure}[ht!]
\begin{subfigure}[b]{0.49\columnwidth}
  \centering
  \includegraphics[width=\linewidth]{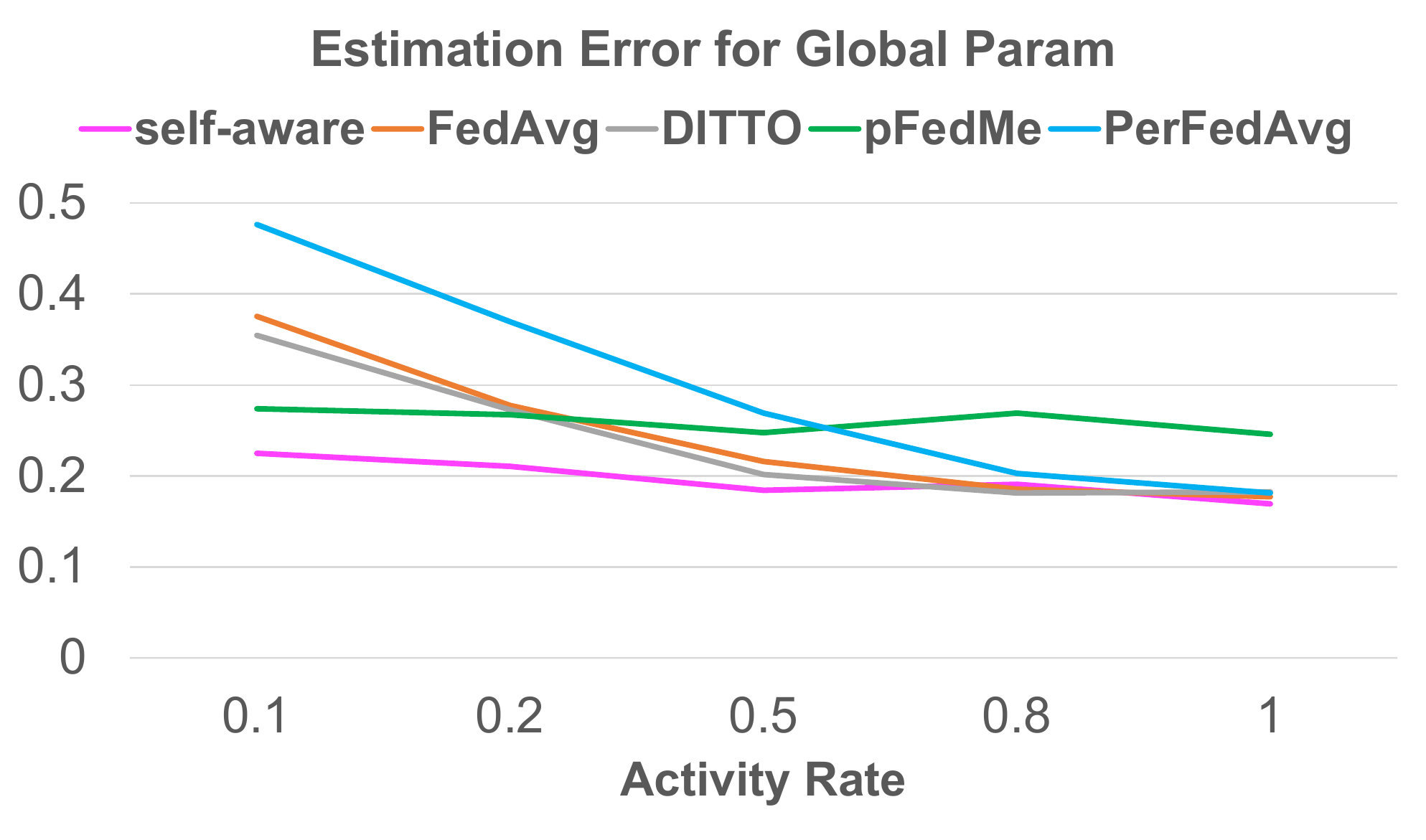}  
  \caption{}
  \label{fig:herero_global}
\end{subfigure}
\begin{subfigure}[b]{0.49\columnwidth}
  \centering
  \includegraphics[width=\linewidth]{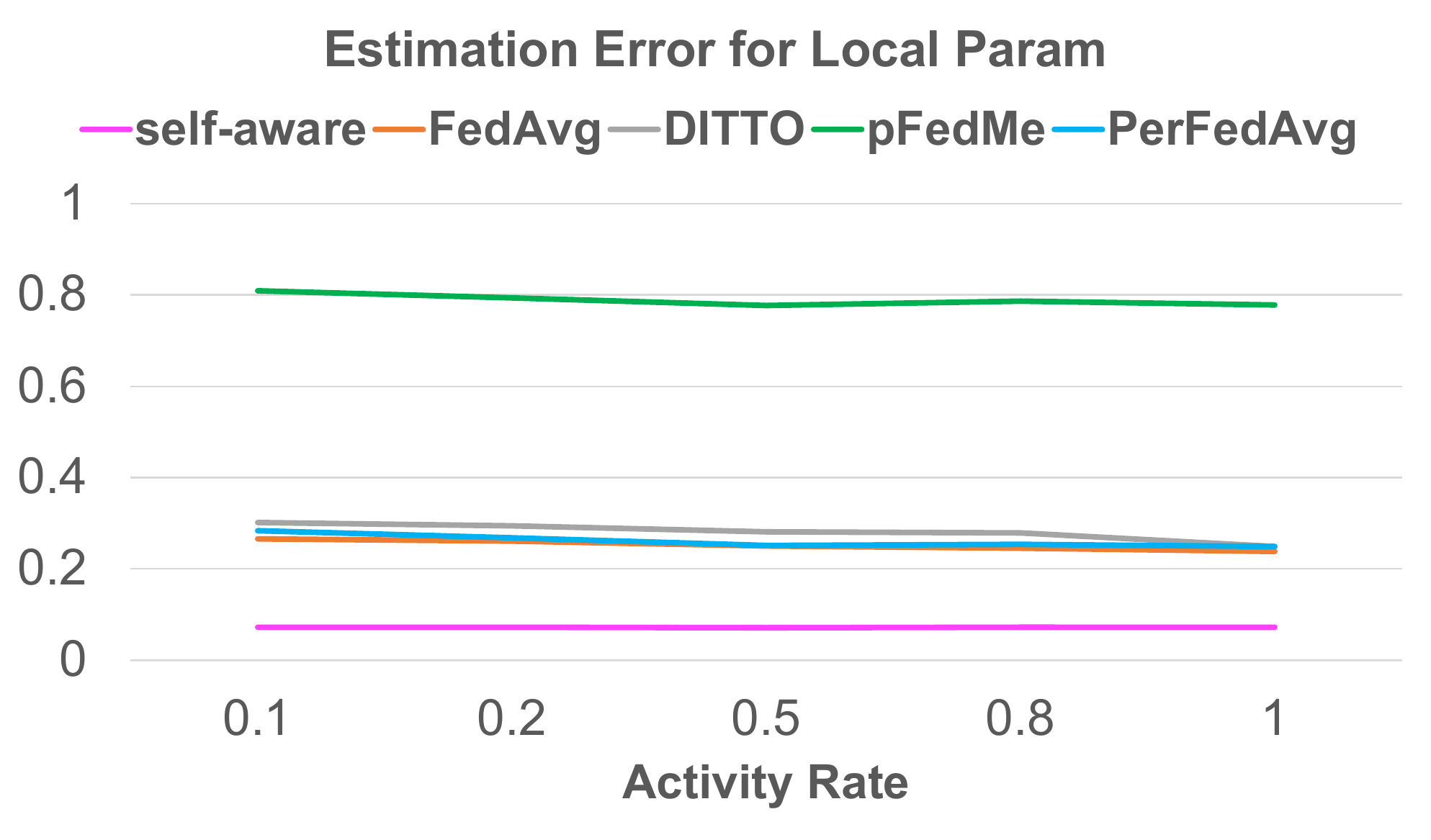}  
  \caption{}
  \label{fig:herero_local}
\end{subfigure}
\vspace{-0.3cm}
\caption{Performance comparison on synthetic datasets of high heterogeneity. Estimation errors are shown for global parameter (a) and local parameter (b). }
\label{fig:comparison_heter}
\end{figure}

Recall that if the inter-client uncertainty is larger, the clients' data are more irrelevant, and the FL posterior approaches the local posterior. Then, the client learns on its own. The global model becomes a simple equal-weighted averaging of local models (detailed in Remark~\ref{remark:interpret_post}). 
We show the performance comparison results of different FL algorithms on the heterogeneous dataset in Figure~\ref{fig:comparison_heter}. One can see from the figure that
\sys{}'s estimation of both the global parameter and the local ones are insensitive to the client participation ratio. 

\vspace{-0.9em}


\subsection{Ablation study of \sys{}'s performance with Algorithms~\ref{algo} and \ref{algo_ablation_complex_version}} \label{app:ablation} 
\vspace{-0.3em}

\begin{figure}[ht!]
\begin{subfigure}[b]{0.45\columnwidth}
  \centering
  \includegraphics[width=\linewidth]{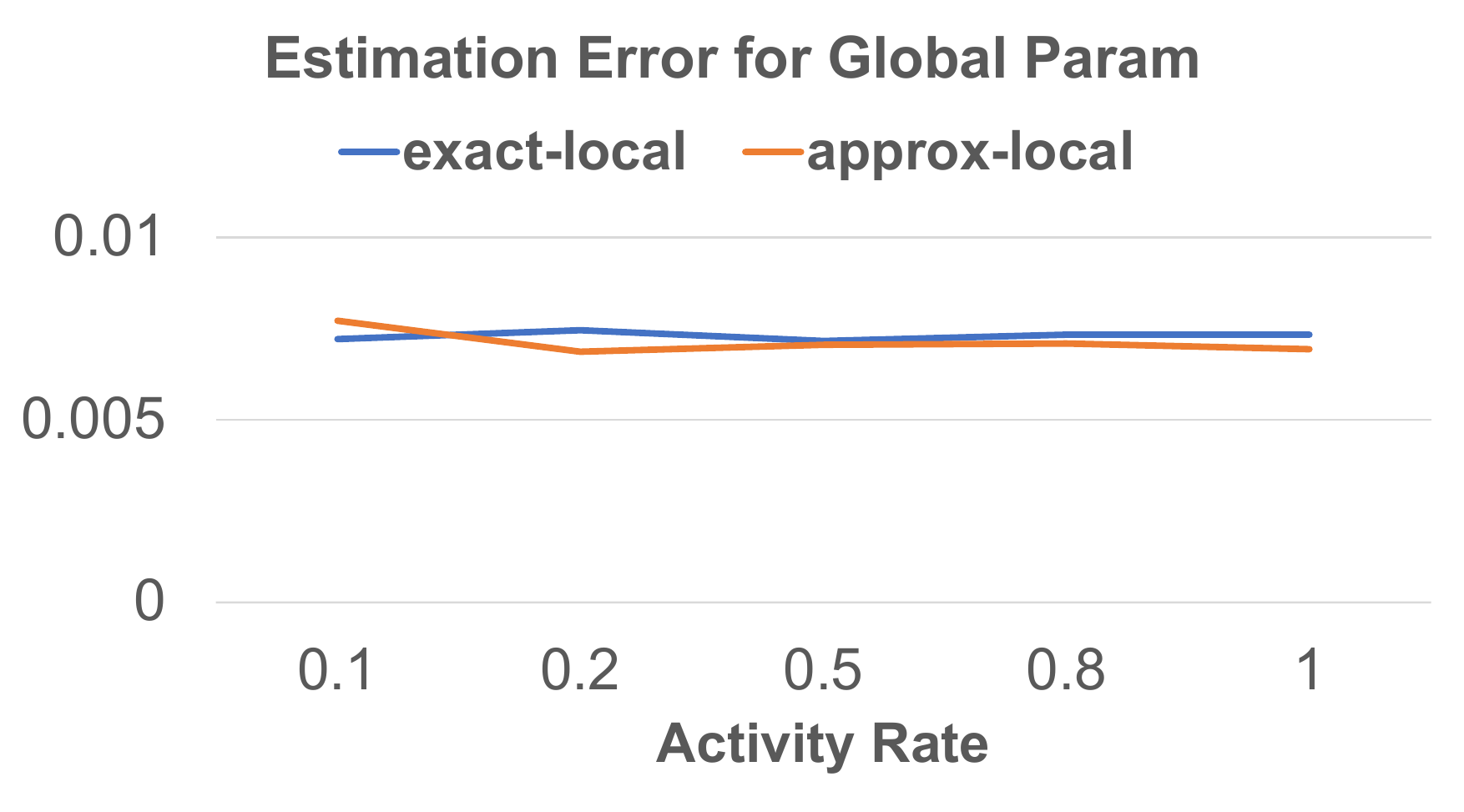}  
  \caption{Homogeneous data}
  \label{fig:ablation_homo}
\end{subfigure}
\hfill 
\begin{subfigure}[b]{0.45\columnwidth}
  \centering
  \includegraphics[width=\linewidth]{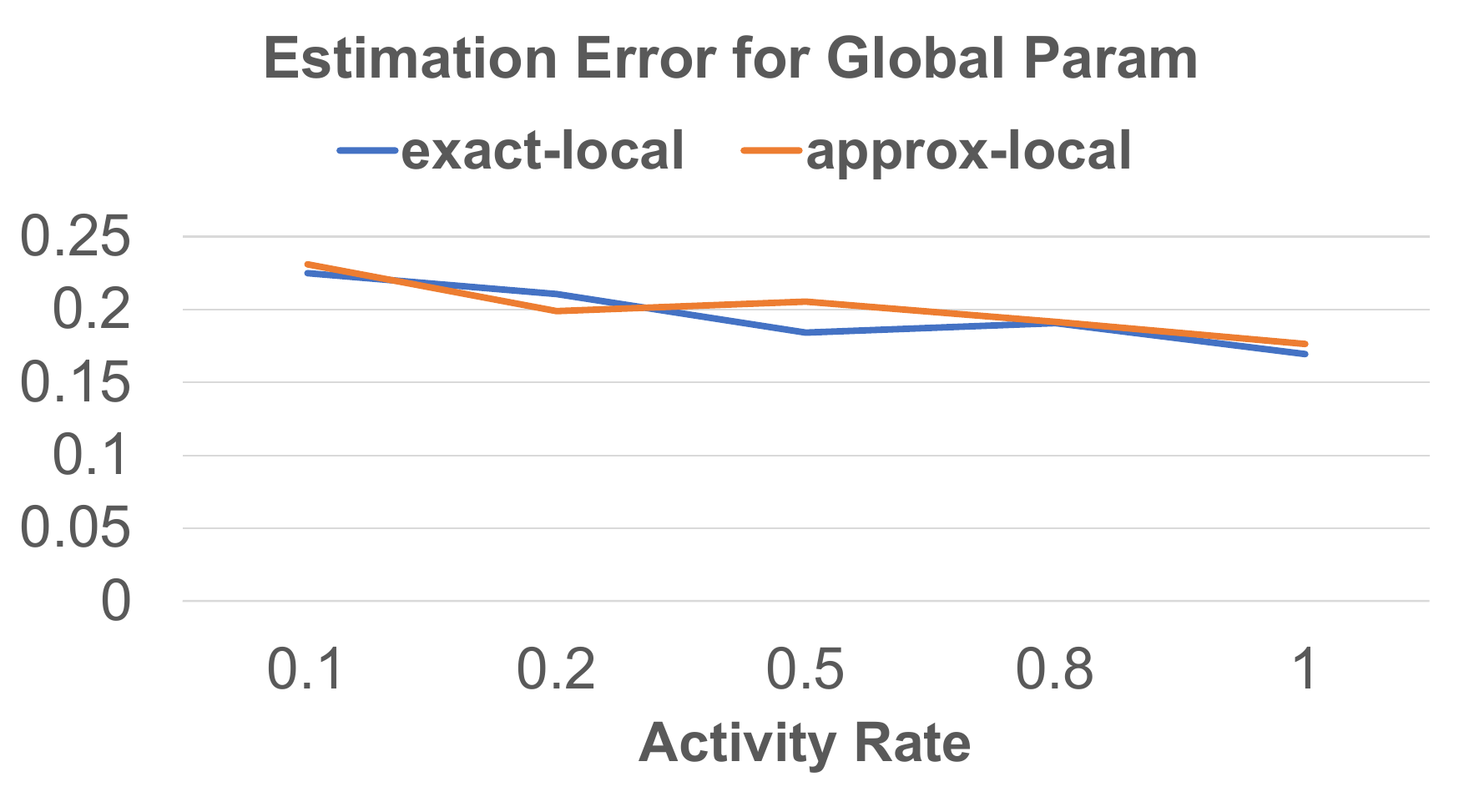}  
  \caption{Heterogeneous data}
  \label{fig:ablation_hetero}
\end{subfigure}
\vspace{-0.7em}
\caption{Performance comparison of Algorithms~\ref{algo} and \ref{algo_ablation_complex_version} on synthetic data.}
\label{fig:ablation_synthetic}
\end{figure}

\noindent \textbf{Results on synthetic data.} In this experiment, we conduct an ablation study of \sys{}'s performance with two different implementation variants outlined in Algorithm~\ref{algo} (using the early-stop parameter $\th_m^t$) and Algorithm~\ref{algo_ablation_complex_version} (using the sufficiently trained local parameter $\th_{m,t}$).   
We use the same synthetic datasets and learning parameters as Section~\ref{sec_exp}. We repeat the experiments for $200$ times and report the mean value. The standard error is kept within $0.001$. 
The results are visualized in Figure~\ref{fig:ablation_synthetic}. 
One can see from the comparison that the performance gap between these two variants is negligible on both homogeneous and heterogeneous data. 

\vspace{-1em}

\begin{figure}[ht!]
\begin{subfigure}[b]{0.40\columnwidth}
  \centering
  \includegraphics[width=\linewidth]{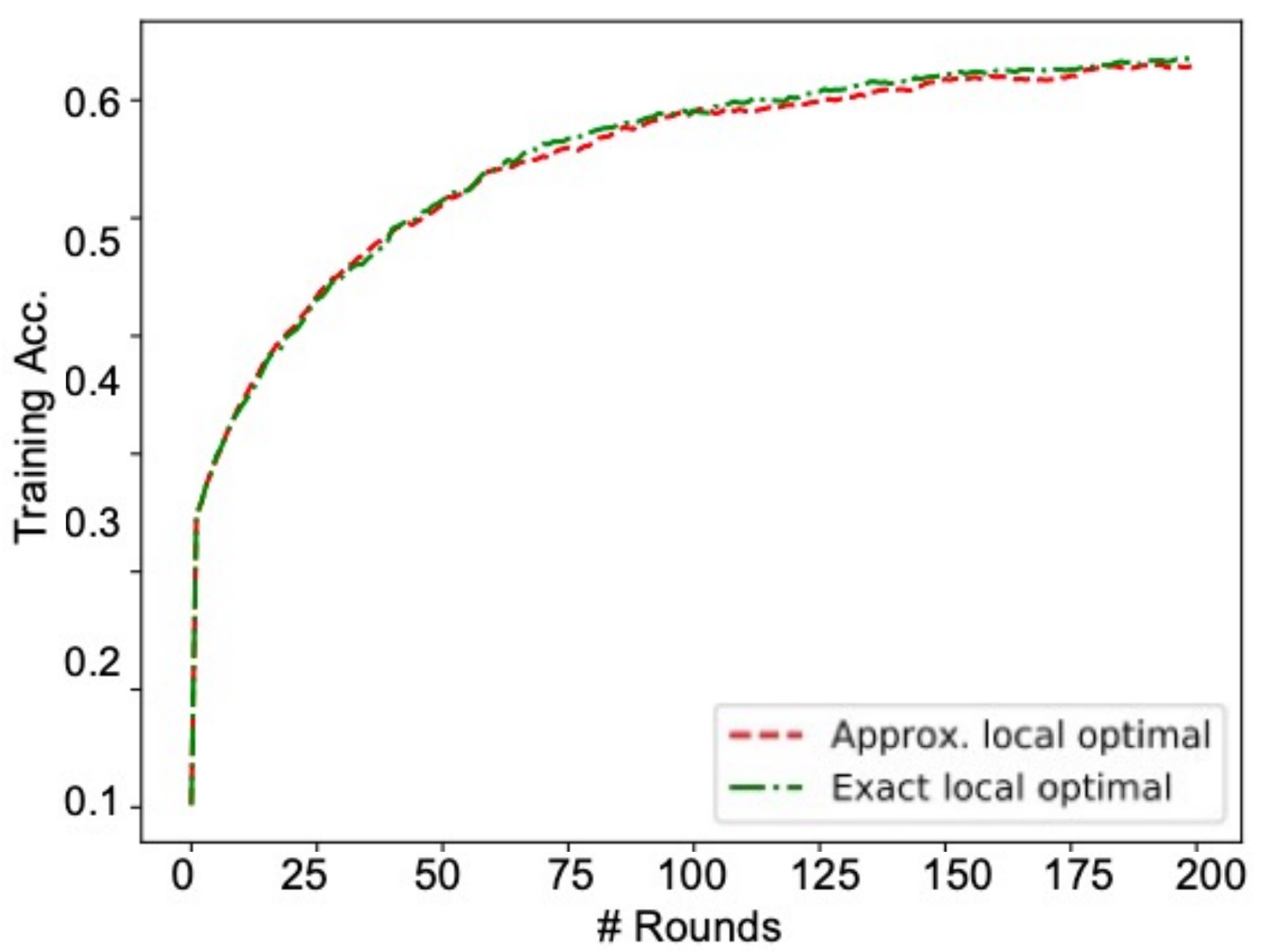}  
  \caption{MNIST}
  \label{fig:ablation_mnist}
\end{subfigure}
\hfill
\begin{subfigure}[b]{0.40\columnwidth}
  \centering
  \includegraphics[width=\linewidth]{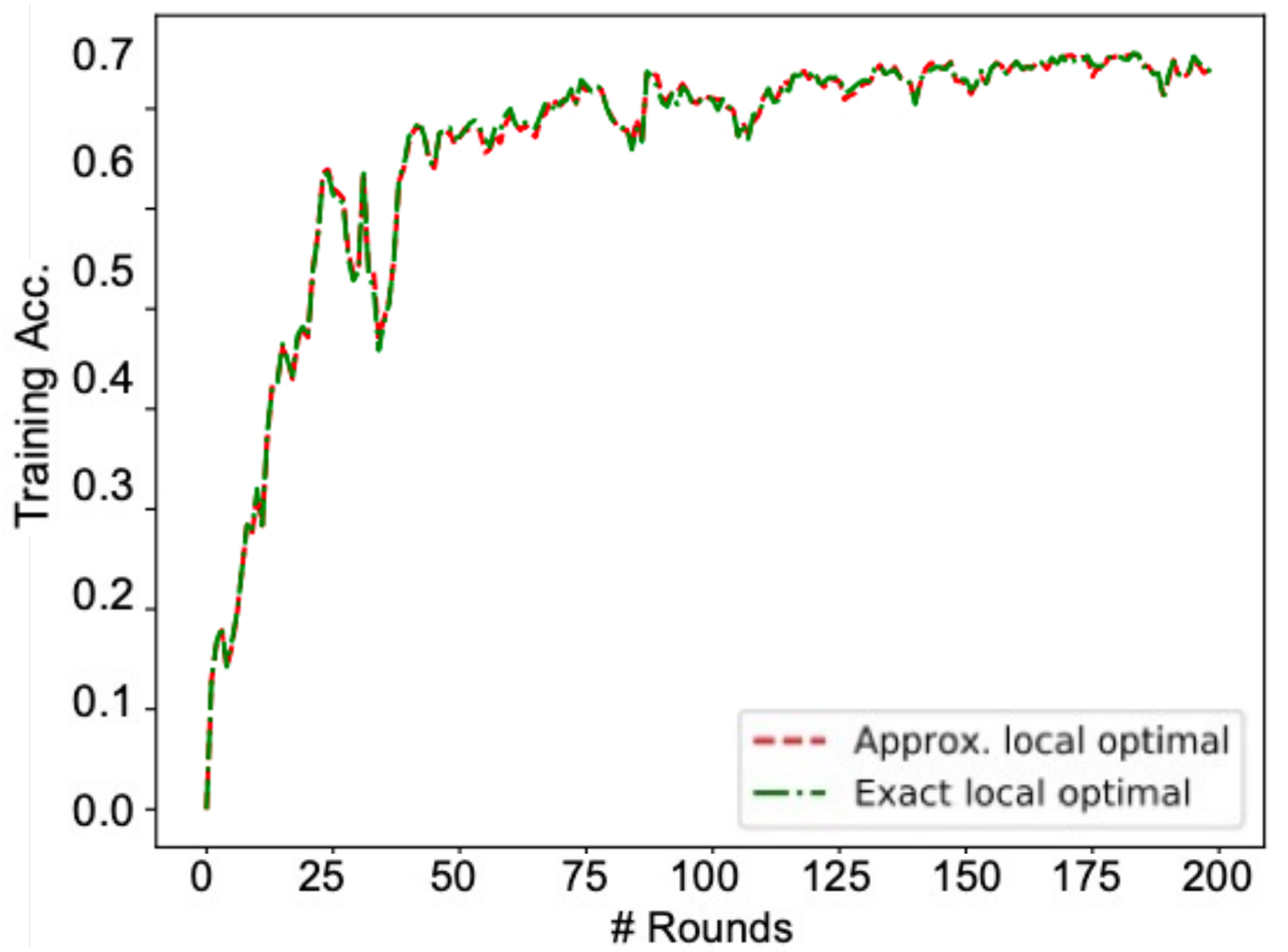}  
  \caption{FEMNIST}
  \label{fig:ablation_femnist}
\end{subfigure}
\vspace{-0.7em}
\caption{Performance comparison of Algorithms~\ref{algo} and \ref{algo_ablation_complex_version} on the MNIST and FEMNIST image data.}
\label{fig:ablation_img}
\end{figure}

\vspace{0.6em}
\noindent \textbf{Results on MNIST and FEMNIST images.} 
Figure~\ref{fig:ablation_img} shows the evaluation results, where the \textit{y}-axis denotes the weighted training accuracy of all personalized models. 
The two curves in Figure~\ref{fig:ablation_femnist} are very close and the accuracy difference is small than $1\%$.  

\vspace{-0.3em}
\noindent \textbf{Results on the Amazon Alexa audio dataset.}
We also perform ablation experiments to assess the performance of two versions of \sys{} on the wake-word audio dataset. The results in Table~\ref{tab:ww_full_ablation} show that their performance is similar.

\begin{table}[ht!]
\centering
\caption{Detection performance (relative FA) of the \textit{global model} on the test dataset of wake-word audio data. The devices are in the normal state. \label{tab:ww_full_ablation} }
\vspace{0.5em}
\scalebox{0.85}{
\begin{tabular}{cccccc} 
\toprule
\multirow{2}{*}{\textbf{FL methods}} & \multicolumn{5}{c}{\textbf{Device Types}} \\
 & \textbf{A} & \textbf{B} & \textbf{C} & \textbf{D} & \textbf{E} \\ 
\midrule
\textbf{Self-FL (Algorithm~\ref{algo_ablation_complex_version})} & 0.93 & 0.94 & 0.96 & 0.89 & 0.98 \\
\textbf{Self-FL (Algorithm~\ref{algo})} & \multicolumn{1}{l}{0.92} & \multicolumn{1}{l}{0.94} & \multicolumn{1}{l}{0.91} & \multicolumn{1}{l}{0.91} & \multicolumn{1}{l}{1.01} \\
\bottomrule
\end{tabular}
}
\end{table}

\subsection{Experiments on Sent140 Dataset}
\label{appendix:eval_sent140}
Sent140~\cite{go2009twitter} is a text sentiment classification dataset. We generate non-i.i.d. FL training data of Sent140 following the same procedure as FedProx~\cite{fedprox}. As for the ML model, we use a two-layer LSTM binary classifier for this experiment. The model has two LSTM layers with 256 hidden units, and the last layer is a fully-connected layer. We use the pre-trained 300D GloVe embedding~\cite{pennington2014glove} to transform the input sequence of 25 characters into the embedding space.
For each FL algorithm, we train the local models with a learning rate of $0.3$, a batch size of $10$, and randomly sub-sample $77$ clients in each communication round (corresponding to a client activity rate $C=0.1$). Each selected local client trains his model for $2$ epochs. This hyper-parameter configuration is suggested in the previous paper~\cite{li2018federated}. Since the number of local epochs is small, we skip \sys{}'s computation of the local training steps $l_m$ and use the same local epochs $l_m=2$.

\vspace{-1em}
\begin{figure}[ht!]
\begin{subfigure}[b]{0.43\columnwidth}
  \centering
  \includegraphics[width=\linewidth]{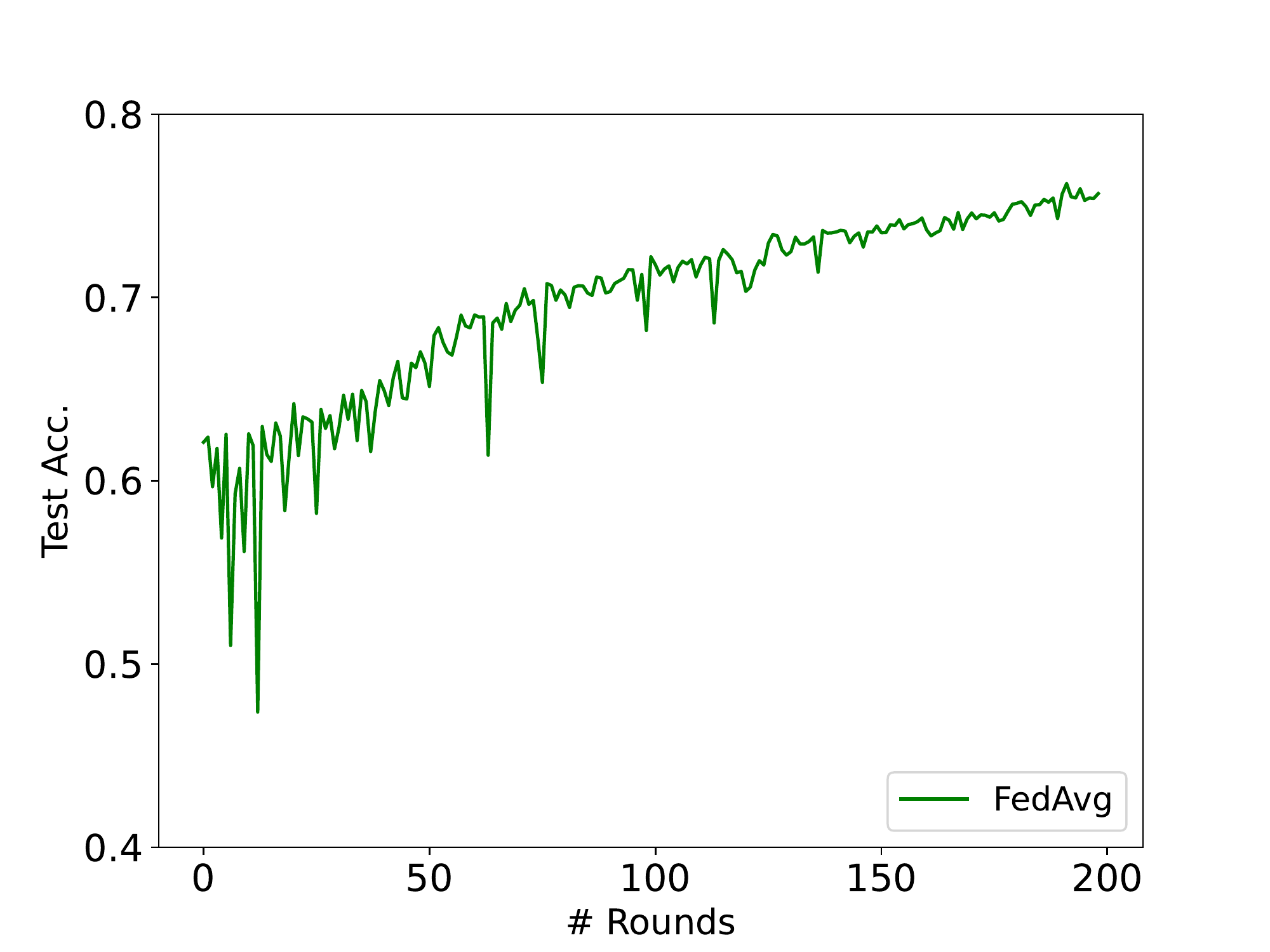}  
  \caption{Training accuracy.}
  \label{fig:sent140_train_warmStart}
\end{subfigure}
\hfill
\begin{subfigure}[b]{0.43\columnwidth}
  \centering
  \includegraphics[width=\linewidth]{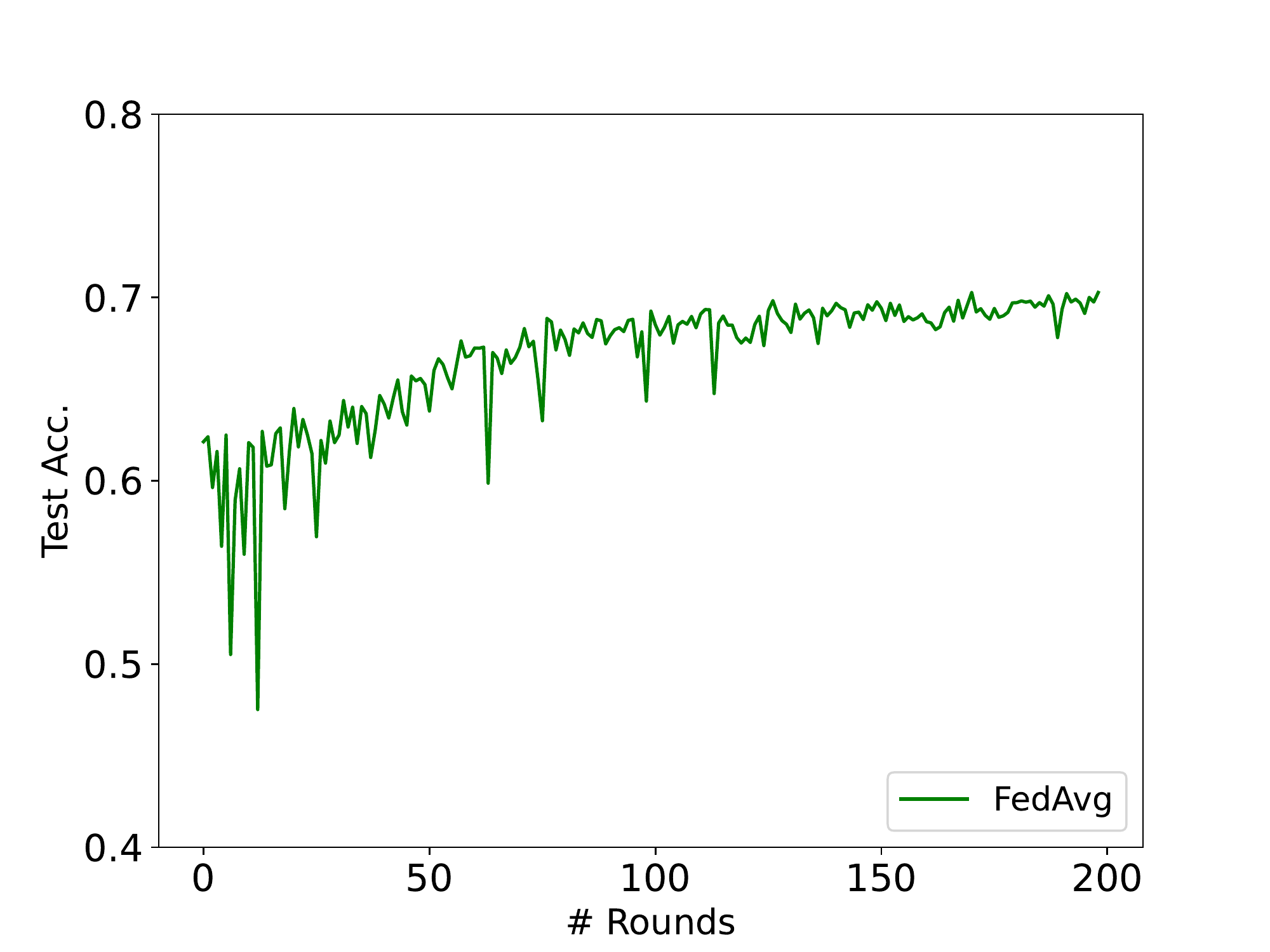}  
  \caption{Test accuracy.}
  \label{fig:sent140_test_warmStart}
\end{subfigure}
\vspace{-0.9em}
\caption{Performance of FL training from scratch using FedAvg (\textit{warm-start} stage). }
\label{fig:sent140_warmStart}
\end{figure}

\vspace{-0.6em}
We use \textit{warm start} in the Sent140 experiment. Particularly, we train a global model from scratch with FedAvg for $200$ rounds and use it as the initialization of the global model for all the other FL algorithms. With this warm start, we continue training the global model with various FL algorithms for another $400$ rounds. 
Figure~\ref{fig:sent140_warmStart} shows the training and test accuracy of the local model obtained by FedAvg~\cite{mcmahan2017communication}. We report the weighted accuracy across all users in the FL system where the weight ratio of each user is proportional to his local sample size.

\begin{figure}[ht!]
\begin{subfigure}[b]{0.46\columnwidth}
  \centering
  \includegraphics[width=\linewidth]{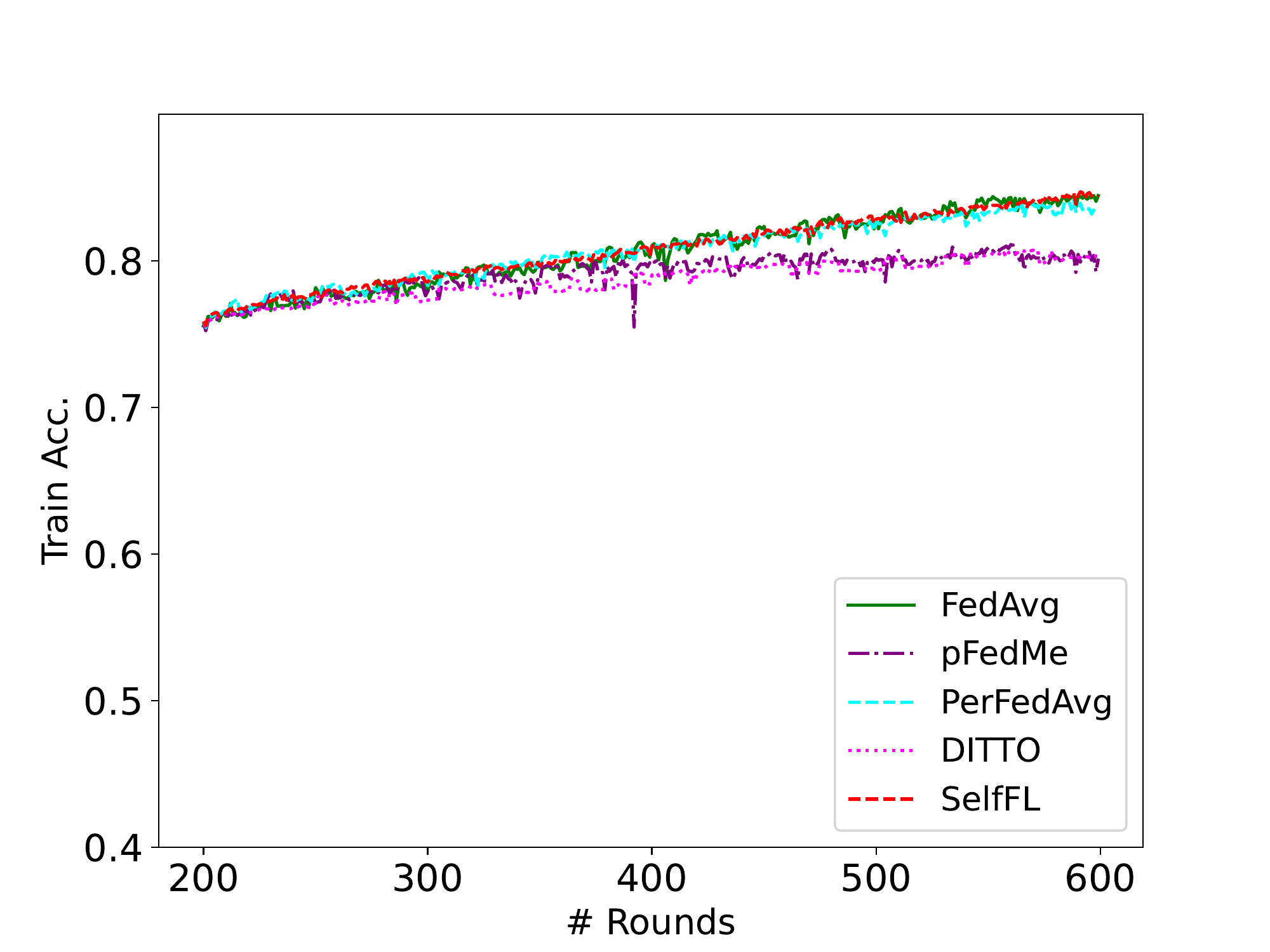}  
  \caption{Training accuracy.}
  \label{fig:sent140_train}
\end{subfigure}
\hfill
\begin{subfigure}[b]{0.46\columnwidth}
  \centering
  \includegraphics[width=\linewidth]{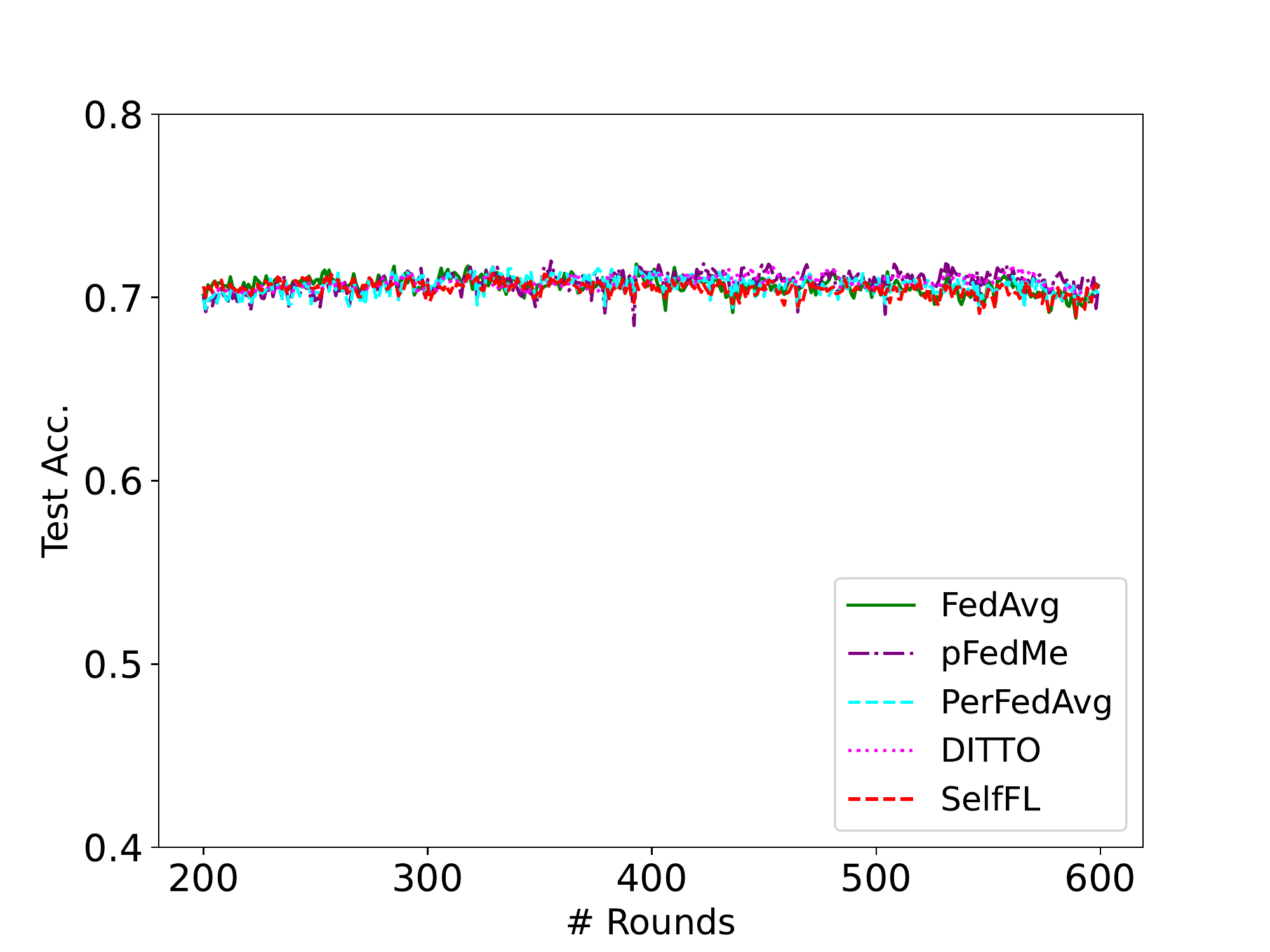}  
  \caption{Test accuracy.}
  \label{fig:sent140_test}
\end{subfigure}
\vspace{-1em}
\caption{Performance comparison of different FL algorithms on Sent140 text data. Note that the x-axis starts from round $200$ since we use warm-start in this experiment. }
\label{fig:sent140_compare}
\end{figure}

With the pre-trained global model from FedAvg (at round $200$) as the `warm-start' initialization, we continue FL training with different algorithms for another $400$ rounds.
Figure~\ref{fig:sent140_compare} compares the training and test accuracy of the personalized/local models obtained by different FL algorithms. The accuracy is aggregated across clients where the aggregation weight is proportional to the local sample size.
Note that Figure~\ref{fig:sent140_compare} shows the accuracy change from round $200$ (where warm start ends) to round $600$. We can see from Figure~\ref{fig:sent140_train} that both \sys{} and FedAvg demonstrates better convergence performance compared to DITTO~\cite{li2021ditto}, pFedMe~\cite{pfedme}, and PerFedAvg~\cite{perfedavg}. 
Recall that the first $200$ rounds are trained with FedAvg and continued by other FL algorithms. As the result, the test accuracy in Figure~\ref{fig:sent140_test} does not change much from round $200$ to round $600$ since it `saturates' during the warm-start stage (Figure~\ref{fig:sent140_test_warmStart}). 

\vspace{-1.2em}

\vspace{0.5em}
\subsection{\sys{}'s performance on wake-word data in clients selection setting}   \label{app:ww_sampling}
\vspace{-0.3em}
In this experiment, we evaluate the performance of FL algorithms in a client selection scenario. More specifically, we assume there are $4$ clients for each of the five device types (A$\sim$E). Each client has a uniform partition of the training set for a specific device type. The FL system has a total of 20 clients. We set the activity rate to $C=0.25$. 

\vspace{-0.3em}
We consider two variants of \sys{} in the clients selection scenario: (i) Client-level variant. This implementation aggregates local models from the currently active clients in each round. (ii) Cluster-level variant. This method groups clients into clusters and then perform aggregation at the cluster level. In our experiments, we cluster clients using device types. We can also use the intra-client uncertainty ($\sigma^2_m$) as the clustering criteria since this information is available for the server (Algorithm~\ref{algo}). With the technique of clients clustering, the server stores the latest model parameter for each cluster as its `representative'. In each communication round, the server loads the weight for each non-active cluster. The global model is updated by aggregating across all clusters. It is worth noting that storing the latest model for each cluster is much more scalable than saving the model separately for each client. 

\vspace{-0.6em}
\begin{table}[b!]
\centering
\caption{Detection performance (relative FA) of the global model obtained using two variants of \sys{} on test data. 
\label{tab:selfFL_variants} }
\vspace{0.5em}
\scalebox{0.8}{
\begin{tabular}{ccccccc} 
\toprule
\multirow{2}{*}{\textbf{FL methods}} & \multirow{2}{*}{\textbf{State}} & \multicolumn{5}{c}{\textbf{Device Type}} \\
 &  & \textbf{A} & \textbf{B} & \textbf{C} & \textbf{D} & \textbf{E} \\ 
\midrule
\multirow{3}{*}{\textbf{\sys{} (client-level)}} & normal & 0.91 & 0.92 & 0.97 & 0.91 & 0.99 \\
 & playback & 0.93 & 0.96 & 1.80 & 1.00 & 1.00 \\
 & alarm & NA & 0.89 & NA & 1.04 & 1.00 \\ 
\midrule
\multirow{3}{*}{\textbf{\sys{} (cluster-level)}} & normal & 0.96 & 0.95 & 1.00 & 0.91 & 0.97 \\
 & playback & 0.86 & 0.99 & 1.00 & 0.97 & 1.01 \\
 & alarm & NA & 0.89 & NA & 1.00 & 1.00 \\
\bottomrule
\end{tabular}
}
\end{table}

The performance of the above two variants is shown in Table~\ref{tab:selfFL_variants}. We can see that the cluster-level aggregation obtains more conservative global models, and its performance improvement is more stable across different device types and operation states compared to the client-level variant. 
This observation is intuitive, because keeping track of the latest weights for each cluster representative for aggregation can prevent the global model from catastrophic forgetting when only a small subset of clients are active in each round.

\subsection{Impact of the activity rate $C$}  \label{app:activity_rate}

\begin{figure}[ht!]
    \centering
    \includegraphics[width=0.45\textwidth]{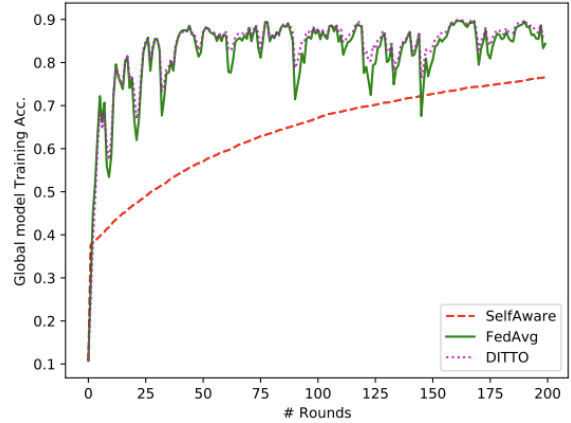}
    \vspace{-1em}
    \caption{Convergence performance of different FL algorithms on the MNIST dataset when the activity rate is set to $0.02$.  }
    \label{fig:mnist_0.02}
\end{figure}

We use the client activity rate as the weight for smoothing the average in the above experiments. While this yields a more stable update, it could cause a slow convergence. 
To investigate the impact of the weights in our smoothing average update for the global model, we set the activity rate to $0.02$ and compare the performance of \sys{} with FedAvg and DITTO. Figure~\ref{fig:mnist_0.02} shows the empirical results in this sparse client participation case. One can see that \sys{} converges slower than FedAvg and DITTO when we use the activity rate ($0.02$) as the weight for the current estimation of the global model in smoothing average. This is expected since a small weight for $\hat{\theta^t}$ means that the update of the global model is more conservative. 

\vspace{-0.7em}

\vspace{-0.3em}
We also perform an ablation study on the wake-word dataset to investigate the impact of the activity rate. The results are summarized in Table~\ref{tab:ww_active_rates}. We use the client-level implementation of \sys{} mentioned in Subsection~\ref{app:ww_sampling} in this experiments. one can see that \sys{} framework obtains better wake-word detection performance (i.e., smaller relative FA values) when the activity rate is not too small. 

\vspace{-1em}
\begin{table}[ht!]
\centering
\caption{\sys{}'s performance with two different activity rates on the wake-word dataset. The relative FA values of the updated global models are reported. \label{tab:ww_active_rates} }
\vspace{-0.5em}
\scalebox{0.9}{
\begin{tabular}{ccccccc} 
\toprule
\multirow{2}{*}{\textbf{Activity Rate}} & \multirow{2}{*}{\textbf{State}} & \multicolumn{5}{c}{\textbf{Device Types}} \\
 &  & \textbf{A} & \textbf{B} & \textbf{C} & \textbf{D} & \textbf{E} \\ 
\midrule
\multirow{3}{*}{\textbf{C = 0.25}} & normal & \textbf{0.91} & 0.92 & 0.97 & 0.91 & 1.00 \\
 & playback & 0.93 & 0.96 & 1.80 & 1.00 & 1.00 \\
 & alarm & NA & 0.89 & NA & 1.04 & 1.00 \\ 
\midrule
\multirow{3}{*}{\textbf{C = 0.1}} & normal & 0.95 & 0.95 & 1.00 & 0.91 & 0.97 \\
 & playback & 0.86 & 0.97 & 1.00 & 0.96 & 1.01 \\
 & alarm & NA & 0.89 & NA & 1.00 & 1.00 \\
\bottomrule
\end{tabular}
}
\end{table}

\subsection{Study of \sys{}'s adaptive local training}  
\label{app:eval_theoretical_lm}
\begin{figure}[ht!]
\begin{subfigure}[b]{0.40\columnwidth}
  \centering
  \includegraphics[width=\linewidth]{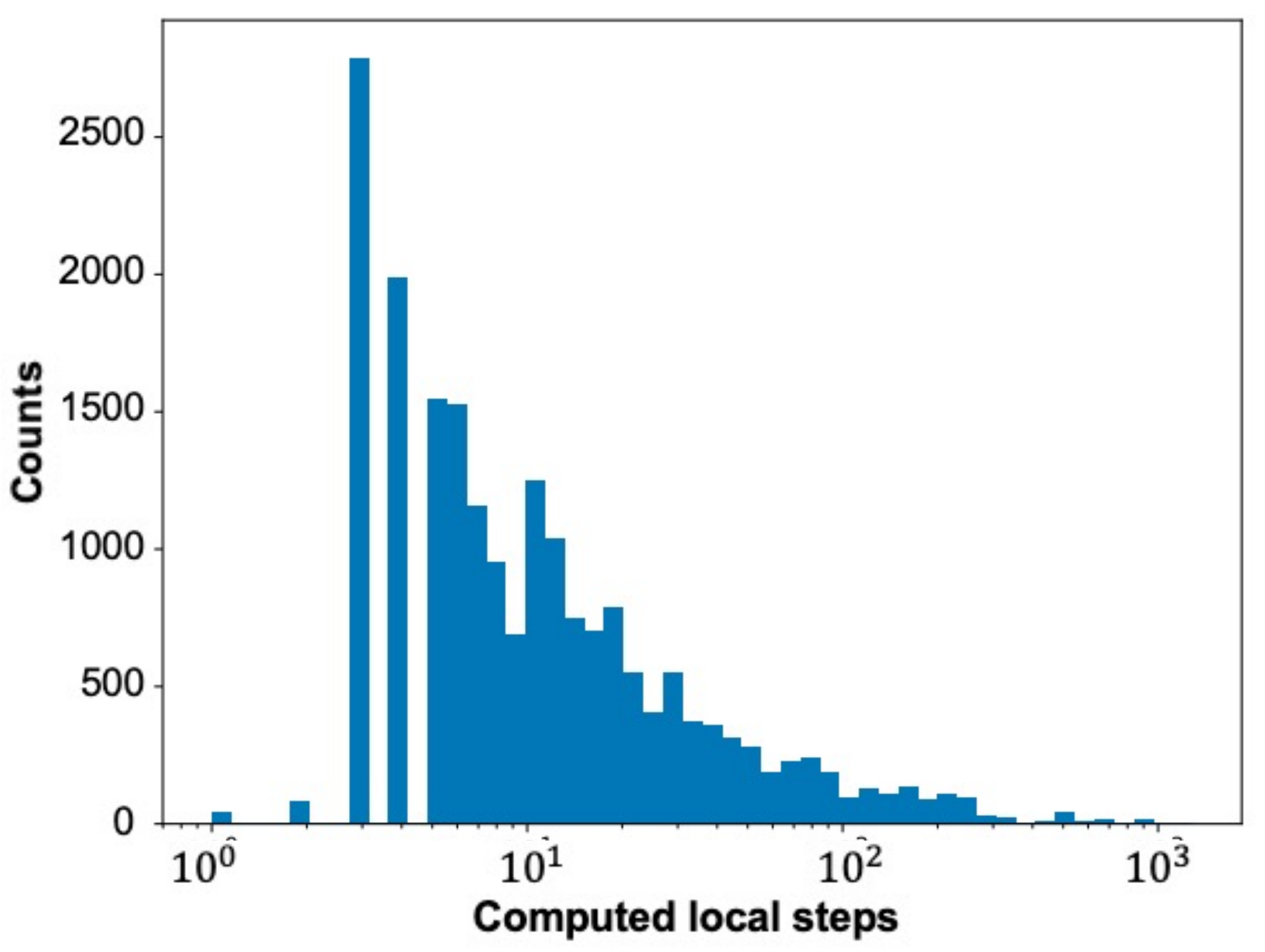}  
  \caption{MNIST}
  \label{fig:mnist_lm_hist_round}
\end{subfigure}
\hfill
\begin{subfigure}[b]{0.40\columnwidth}
  \centering
    \includegraphics[width=\linewidth]{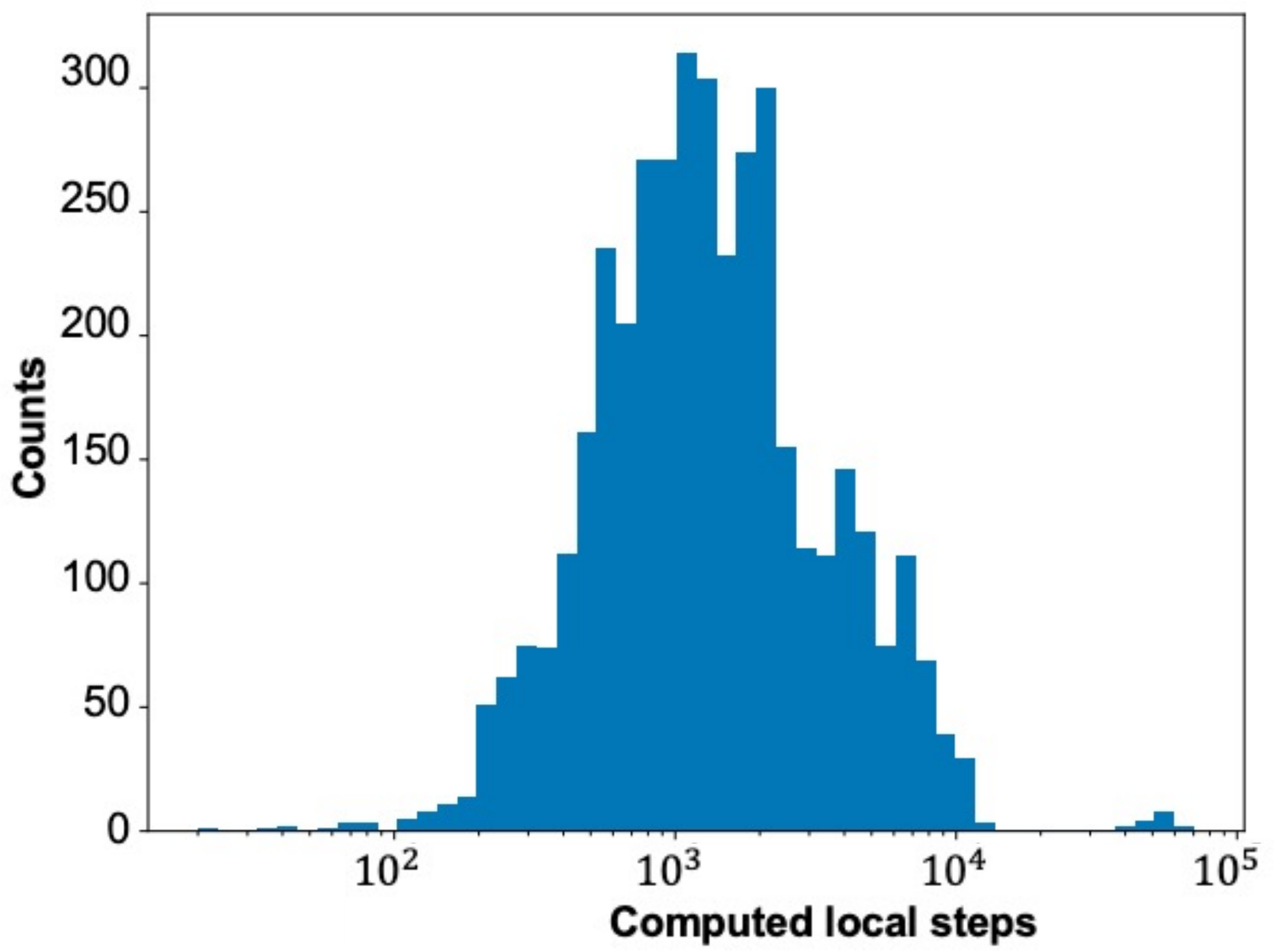}    
  \caption{FEMNIST}
  \label{fig:nist_lm_hist_round}
\end{subfigure}
\vspace{-0.8em}
\caption{Histogram of the computed number of local training steps for active clients at the round $t=100$. }
\label{fig:lm_hist}
\end{figure}

\begin{figure}[ht!]
\begin{subfigure}[b]{0.40\columnwidth}
  \centering
  \includegraphics[width=\linewidth]{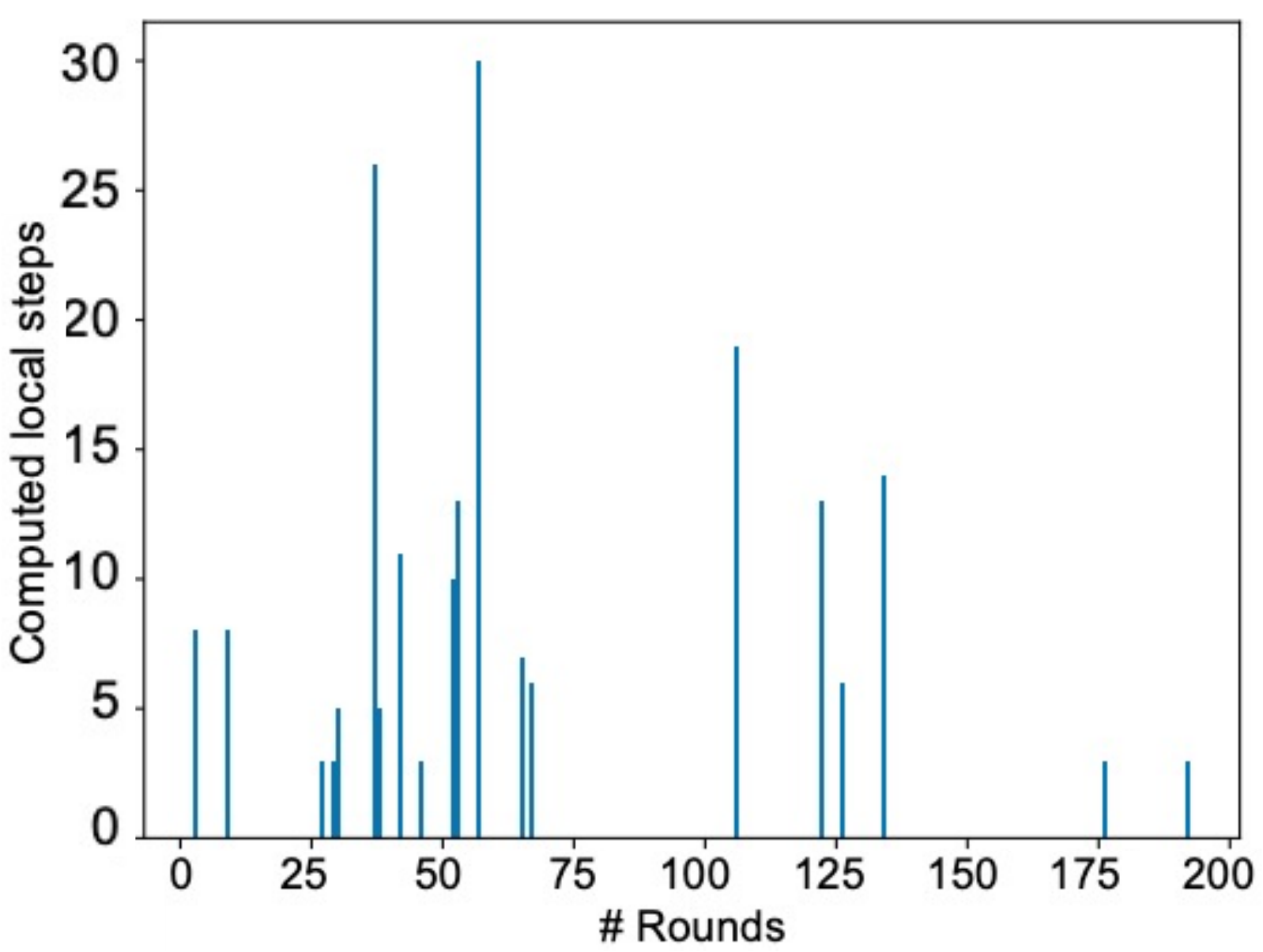}  
  \caption{MNIST}
  \label{fig:mnist_lm_cli10_allrounds}
\end{subfigure}
\hfill 
\begin{subfigure}[b]{0.40\columnwidth}
  \centering
  \includegraphics[width=\linewidth]{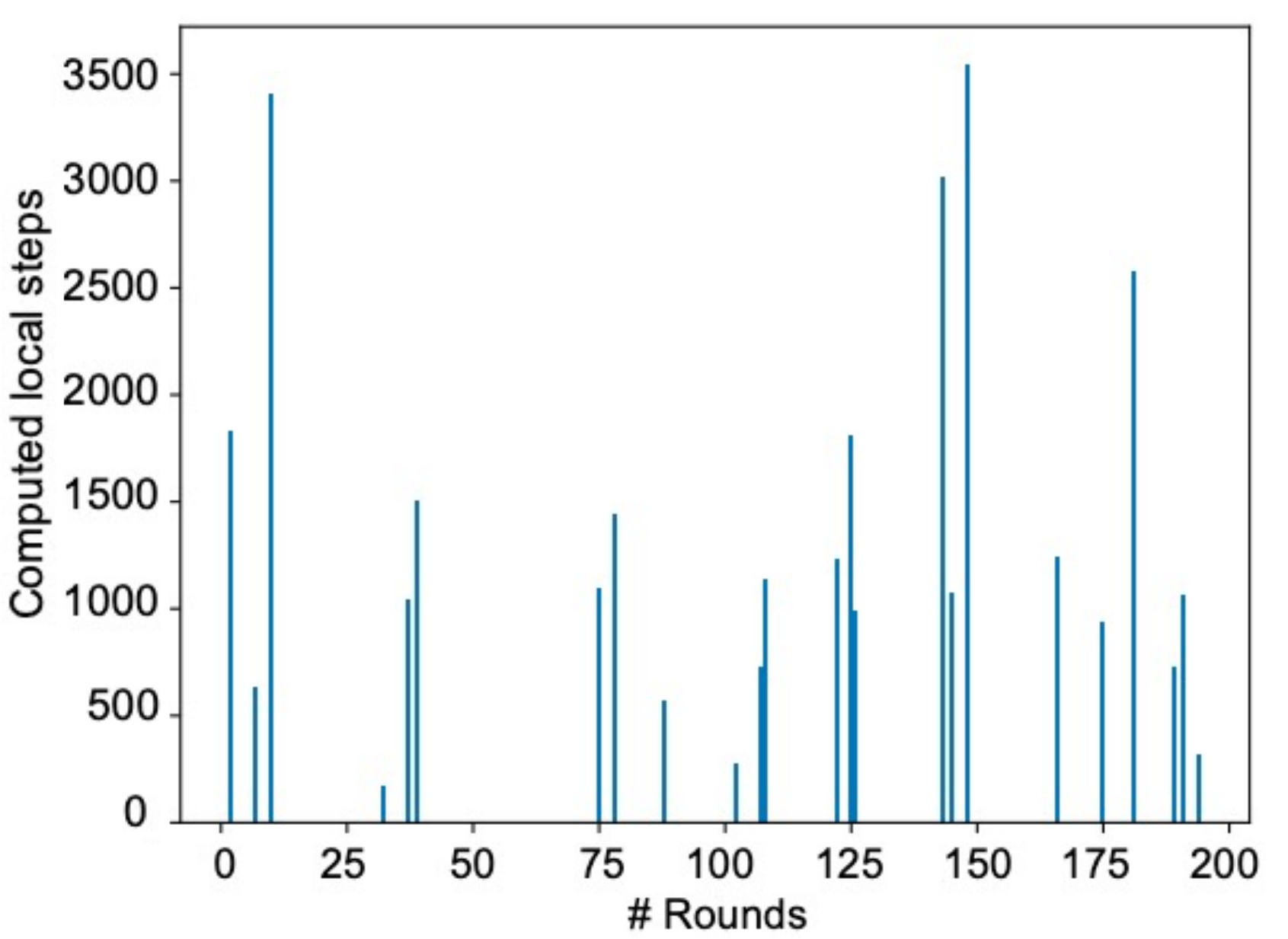}  
  \caption{FEMNIST}
  \label{fig:nist_lm_cli2_allrounds}
\end{subfigure}
\vspace{-0.5em}
\caption{Trajectory of the computed local training steps for a randomly selected client in different FL rounds. 
}
\label{fig:variation_lm_rounds}
\end{figure}

A key component of \sys{} is the data-adaptive local training procedure as discussed in Section~\ref{sec_method}. 
The local optimization trajectory is designed in such a way that it achieves a good balance between local model improvement and global model update. 
We study the theoretical values of local training steps $l_m$ computed using Eqn.~(\ref{eq:lm}). 
Figure~\ref{fig:lm_hist} shows the histogram of $l_m$ of the active clients in round $t=100$. 
We also visualize the dynamics of the local training steps for a randomly selected client across FL rounds in Figure~\ref{fig:variation_lm_rounds}. The value of $l_m$ exists for a subset of rounds since we use the activity rate $C=0.1$. We can see that the computed number of steps $l_m$ has an overall decreasing trend after round $150$, suggesting a drop of the intra-client uncertainty $\sigma_m^2$. In other words, the quality of personalized model $\theta^t_m$ improves eventually.  

\section{Proof of Proposition~\ref{prop_converge}}  
\label{app:proof_proposition}
In this section, we provide detailed proof of Proposition~\ref{prop_converge} introduced in the main paper. 
For brevity, we write $\sum_{k \in [M]}$ and $\sum_{k \in [M]-\{m\}}$ as $\sum_{k}$ and $\sum_{k:k\neq m}$, respectively. We define $\sum_{i,j: \, i\neq j}$ similarly.

Let $w_m = (\vD_m + \vT)^{-1}$.
According to the assumption, let $\cv$ be an upper bound of $\max_{m\in [M]} \vD_m$. Then 
we have
\begin{align}
    \wmin \de (\cv + \vT)^{-1} \leq \min_{m \in [M]} w_m 
    \leq \max_{m \in [M]} w_m \leq \ivT \de \wmax . \label{eq_w}
\end{align}
Let $v = \max_{m\in [M]} (w_m/\sumkA w_k)$. It follows from (\ref{eq_w}) that 
\begin{align}
    v=O(M^{-1}) \label{eq202}
\end{align}
for large $M$.
Let 
\begin{align}
    \zeta^{t}_m \de \hat{\th}^{t}_m - \thFL_m , \quad \forall m \in [M], t =0, 1, \ldots  \label{eq204}
\end{align}
Recall that by the assumption, $C$ is a constant upper bound of $\max_{m \in [M]} |\zeta^{0}_m|$.
Next, we will provide a uniform bound on $|\zeta^{t}_m|$, by considering every $t \geq 1$.

According to the update rule, as shown in equation (\ref{eq2}), we have
\begin{align}
    \th^{t}_m = \frac{\ivD_m }{\ivD_m + \sumK } \thL_m + \frac{\sumK}{\ivD_m + \sumK}  \th^{t,\init}_m , \label{eq100} 
\end{align}
where $\th^{t,\init}_m $ is the initial parameter of client $m$ at time $t$, introduced in (\ref{eq_101}).
Recall from (\ref{eq1}) that 
\begin{align}
    \thFL_m = \frac{\ivD_m \thL_m + \sumK \thL_k}{\ivD_m + \sumK } .\label{eq102}
\end{align}
Combining (\ref{eq100}) and (\ref{eq102}), and invoking the definition in (\ref{eq103}), we have
\begin{align}
    |\zeta^{t}_m| = |\th^{t}_m - \thFL_m|
    &= \biggl|\frac{\sumK}{\ivD_m + \sumK } \biggl(\th^{t,\init}_m - \frac{\sumK \thL_k}{ \sumK } \biggr) \biggr| \\
    &\leq q \biggl|\th^{t,\init}_m - \frac{\sumK \thL_k}{ \sumK } \biggr| \label{eq206}
\end{align}

By the definition of $\th^{t,\init}_m $, we may rewrite it as 
\begin{align}
    \th^{t,\init}_m 
    = \th^{t-1} -  \frac{(\vT + \vD_m)^{-1} }{\sum_{k: \, k \neq m} (\vT + \vD_k)^{-1}} (\th^{t-1}_m - \th^{t-1})
    = \frac{\sumK \th^{t-1}_m}{\sumK} .
\end{align}
Therefore, with the definition in (\ref{eq204}), we have 
\begin{align}
    \th^{t,\init}_m - \frac{\sumK \thL_k}{ \sumK } 
    &= \frac{\sumK (\th^{t-1}_k - \thL_k )}{\sumK} \\
    &= \frac{\sumK (\zeta^{t-1}_k + \thFL_k  - \thL_k )}{\sumK} \\
    &= \frac{\sumK \zeta^{t-1}_k}{\sumK} + \frac{\sumK \thFL_k }{\sumK} - \frac{\sumK \thL_k}{\sumK} . \label{eq205}
\end{align}
Meanwhile, it can be verified that
\begin{align}
    &\frac{\sumK \thFL_k }{\sumK} = \frac{\sumkA w_k \thFL_k }{\sumkA w_k} (1-v_m)^{-1} - \frac{v_m}{1-v_m} \thFL_m, 
    \quad \textrm{ where } 
    v_m \de \frac{w_m}{\sumkA w_k} \leq v, \\
    &\frac{\sumK \thL_k }{\sumK} = \frac{\sumkA w_k \thL_k }{\sumkA w_k} (1-v_m)^{-1} - \frac{v_m}{1-v_m} \thL_m, 
\end{align}
which further implies that 
\begin{align}
    &\biggl| \frac{\sumK \thFL_k }{\sumK} - \frac{\sumK \thL_k }{\sumK} \biggr| 
    \leq \frac{1}{1-v} |D| + \frac{2v c_{\theta}}{1-v} , \label{eq207} \\
    &\textrm{ where }  D \de \frac{\sumkA w_k (\thFL_k -\thL_k) }{\sumkA w_k}, \label{eq208}
\end{align}
and $c_{\theta}$ is an upper bound of $|\thFL_m|$ and $|\thL_m|$.
The definition of $\thFL_m$ implies that $|\thFL_k| \leq \max_{m \in [M]} |\thFL_m|$ for all $k \in [M]$.
Since $\thL_m$'s are independent Gaussian with variance bounded by $\cv$, we have
$\max_{m \in [M]} |\thFL_m| = O_p(\sqrt{\log M})$. Thus, we may choose
\begin{align}
    c_{\theta} =  O_p(\sqrt{\log M}). \label{eq401}
\end{align}

Taking (\ref{eq205}) and (\ref{eq207}) into (\ref{eq206}), we obtain  
\begin{align}
    &|\zeta^{t}_m| \leq q \max_{m \in [M]} \biggl|\frac{\sumK \zeta^{t-1}_k}{\sumK}\biggr| + |D|
    \leq q \max_{m \in [M]} |\zeta^{t-1}_m| + \frac{|D|+ 2v c_{\theta}}{1-v}
    , \quad \forall m \in [M]. \label{eq_210}
\end{align}
It follows from (\ref{eq_210}) that 
\begin{align}
    \max_{m \in [M]} |\zeta^{t}_m| \leq q^{t} \max_{m \in [M]} |\zeta^{0}_m| + \frac{|D|+ 2v c_{\theta}}{1-v}. \label{eq210}
\end{align}

Next, we bound $|D|$.
Recall that $\E(\thFL_i) = \th_0$ and $\cov(\thFL_i, \thFL_j)=0$ for all $i, j \in [M] $ and $i \neq j$.
Since
\begin{align}
    D &= \frac{\summA w_m (\thFL_m -\thL_m) }{\summA w_m}, \\
    \thFL_m -\thL_m 
    &= \frac{\sum_{k: \, k\neq m} w_k \thL_k}{\ivD_m + \sumK } - \frac{\sum_{k: \, k \neq m} w_k}{\ivD_m + \sumK } \thL_m 
    = \frac{\sum_{k: \, k \neq m} w_k (\thL_k - \thL_m)}{\ivD_m + \sumK } \\
    \var(\thFL_m -\thL_m ) 
    &\leq 2 \var\biggl(\frac{\sum_{k: \, k \neq m} w_k \thL_k}{\ivD_m + \sumK }\biggr) + 2 \var\biggl(\frac{\sum_{k: \, k \neq m} w_k}{\ivD_m + \sumK } \thL_m \biggr) 
    \leq 4 \max_{k}  \var(\thL_k) = 4 V,
\end{align}
we have $\E(D) = 0$, and 
\begin{align}
    \var(D) 
    &=  
    \frac{\summA w_m^2 \var(\thFL_m -\thL_m) }{(\summA w_m)^2}
    + \frac{\sum_{i,j: \, i\neq j} w_i w_j \cov(\thFL_i -\thL_i, \thFL_j -\thL_j) }{(\summA w_m)^2} \\
    &\leq \frac{4 V \summA w_m^2 }{(\summA w_m)^2} + \sum_{i,j: \, i\neq j}\frac{ w_i w_j \cov(\sum_{k: \, k \neq i} w_k (\thL_k - \thL_i), \sum_{k: \, k \neq j} w_k (\thL_k - \thL_j)) }{(\summA w_m)^2 (\ivD_i + \sum_{k: \, k \neq i} w_k) (\ivD_j + \sum_{k: \, k \neq j} w_k ) } \\
    &= \frac{4 V \summA w_m^2 }{(\summA w_m)^2} + \sum_{i,j: \, i\neq j}w_{i} w_{j} \frac{ \sum_{k: \, k \neq i,j} w_k^2 \var(\thL_k) -  \biggl( w_j\var(\thL_{j})\sum_{k: \, k\neq j} w_k +w_i\var(\thL_{i})\sum_{k: \, k\neq i} w_k \biggr) }{(\summA w_m)^2 (\ivD_i + \sum_{k: \, k \neq i} w_k) (\ivD_j + \sum_{k: \, k \neq j} w_k ) } \\
    &\leq \frac{4 V \summA w_m^2 }{(\summA w_m)^2} + \sum_{i,j,k: \, i\neq j,k\neq i, k\neq j}   \frac{ w_i w_j w_k^2 V }{(\summA w_m)^2 (\ivD_i + \sum_{k: \, k \neq i} w_k) (\ivD_j + \sum_{k: \, k \neq j} w_k ) } \\
    &\leq \frac{4V }{M} \frac{\wmax^2}{\wmin^2} + \frac{M V \wmax^4}{\wmin^2 (\min_{m \in [M]} \ivD_m + (M-1) \wmin)^2 } \\
    &\leq \frac{4V }{M} \frac{\wmax^2}{\wmin^2} + \frac{M V }{(M-1)^2}\frac{\wmax^4}{ \wmin^4 }.
\end{align}
Therefore, from (\ref{eq_w}), we have $\var(D) = O(M^{-1})$ as $M \rightarrow \infty$. By the Markov inequality, we further have 
\begin{align}
    |D| = O_p(M^{-1/2}) .  \label{eq101}
\end{align}
Consequently, taking equations (\ref{eq202}), (\ref{eq401}), and (\ref{eq101}) into (\ref{eq210}), we obtain
\begin{align}
    \max_{m \in [M]} |\zeta^{t}_m| = c_3 \cdot q^{t}  +  O_p(M^{-1/2}) + O(M^{-1})   
    = C \cdot q^{t}  +  O_p(M^{-1/2}) \textrm{ as }  M \rightarrow \infty, 
\end{align}
where $C$ is the constant upper bound of $\max_{m \in [M]} |\zeta^{0}_m|$ (by the assumption).
This proves the convergence of each client's personalized model. 

For the server model denoted by $\th^t$, since
\begin{align}
    \th^t 
    & = \frac{\sumkA w_k \th^t_k }{\sumkA w_k} 
     = \frac{\sumkA w_k (\th^t_k - \thFL_k) }{\sumkA w_k} + \frac{\sumkA w_k (\thFL_k -\thL_k) }{\sumkA w_k} + \frac{\sumkA w_k \thL_k }{\sumkA w_k} \\
    & = \frac{\sumkA w_k \zeta_k }{\sumkA w_k} + D + \thG,
\end{align}
we have
$|\th^t - \thG| \leq C \cdot q^{t} + O_p(M^{-1/2})$. 
This concludes the proof.


\end{document}